%% file: acl_latex.tex
\documentclass[11pt]{article}

\usepackage[final]{acl}

\usepackage{times}
\usepackage{latexsym}

\usepackage[T1]{fontenc}

\usepackage[utf8]{inputenc}

\usepackage{microtype}

\usepackage{inconsolata}

\usepackage{graphicx}

\usepackage{algorithm}
\usepackage{algpseudocode}
\usepackage{xcolor}
\usepackage{booktabs}
\usepackage{listings}
\usepackage{amsmath}
\usepackage{hyperref}

\title{TESS: LLM-guided Translation, Exploration, Simulation, and Search for Embodied Task Planning}

\author{
 \textbf{Peiran Xu\textsuperscript{1}},
 \textbf{Yixiang Yang\textsuperscript{1}},
 \textbf{Jinghan Li\textsuperscript{1}},
 \textbf{Fei Li\textsuperscript{2}},
 \textbf{Cao Sheng\textsuperscript{2}},
 \textbf{JiaZhong Yu\textsuperscript{2}},
 \textbf{Yadong Mu\textsuperscript{1}} \thanks{Correspondence author.}
\\
\\
 \textsuperscript{1}Peking University,
 \textsuperscript{2}China Tower
}

\begin{document}
\maketitle
\begin{abstract}
In this work we focus on the problem of embodied task planning, which requires an agent to accomplish a complex task by separating it into atomic actions. Recent studies have developed LLM-based methods to address this problem. However, the executability and correctness of the generated plans remain suboptimal in complex environments and tasks. On the other hand, classical search algorithms have the potential to solve the planning problems perfectly, yet they require fully observable environments, known state transitions and goal state, which are often unattainable in general task domains. Therefore, we propose to implement an LLM-guided search algorithm that makes the search process possible and efficient by incorporating LLMs' commonsense knowledge. Our method leverages LLMs to (1) translate natural language instructions into formatted goal conditions, (2) facilitate efficient exploration in partially observable environments, (3) implement a rule-based world simulator, and (4) prune the space of candidate actions to improve search efficiency. Experimental results on VirtualHome and ALFWorld demonstrate its superior performance. 
\end{abstract}

\input{sec/1_intro}
\input{sec/2_related}
\input{sec/3_method}

\input{sec/4_exp}
\input{sec/5_conclusion}

\section{Limitations}
This work focuses exclusively on high-level reasoning and planning, while abstracting away visual perception and low-level robotic control. In addition, we assume goal-oriented user instructions and do not consider tasks involving temporal constraints, procedural requirements, or more open-ended instructions. Also, the current planning pipeline does not address dynamically changing environments and non-deterministic actions, it relies on a known action space, and cannot explicitly distinguish an unsolvable problem. 

To the best of our knowledge, this work does not introduce any new risks to the society and research community, aside from the potential for harmful outputs that generally exists in LLM research. Future efforts could focus on generating safe and reliable plans under constrained conditions in order to avoid dangerous or undesirable behaviors, especially when the agent is deployed in real-world environments.

\section*{Acknowledgments}
The research was supported by a joint project between Peking University and China Tower.
\bibliography{custom}

\appendix

\input{sec/X_supp}

\end{document}

%% file: sec/1_intro.tex
\vspace{-4mm}
\section{Introduction}
\label{sec:intro}
Planning is a fundamental capability for general-purpose robots. In the problem of Embodied Task Planning, low-level robotic control is abstracted into a set of atomic skills, and the agent is expected to complete a user-specified task by performing an appropriate sequence of these skills in an interactive environment. This process requires in-depth reasoning over the environment state, user’s instruction, and the history of executed actions. As a crucial step toward achieving embodied intelligence for solving complex tasks, this problem has attracted growing attention, and a number of simulators and benchmarks have been built~\cite{VirtualHome,AW,B1k,EAI}.

With the advancement of large language models (LLMs) in reasoning and planning tasks, many recent studies have explored using LLMs as the core component for embodied task planning. Typically, an LLM is prompted or finetuned to generate a full plan or function as an action policy. Despite the notable progress achieved by previous works through improved training strategies~\cite{LMIDM,LMWM,ToolkenGPT,plasma,ESM} and test-time scaling~\cite{LLMCK,E2CL,TreePlanner,DGEP}, ensuring the executability and correctness of the generated plans remains a major challenge.
The limitation may stem from the fact that LLMs still struggle with reasoning in long contexts (containing the description of a large-scale environment and an extended action history) and planning for complex tasks (analyzing the intent of the user, the dependencies among subtasks, and the feasibility of actions).

\begin{figure*}
    \centering
    \includegraphics[width=0.88\linewidth]{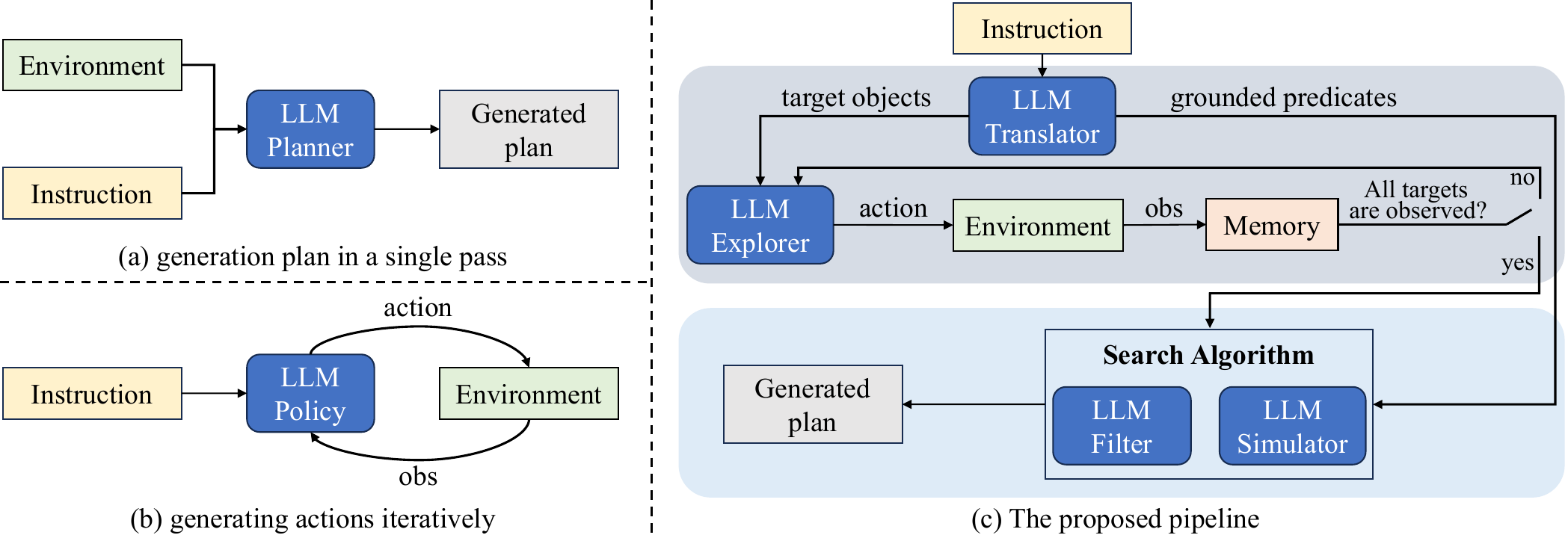}
    \vspace{-1mm}
    \caption{A high-level comparison of (a) LLM-based plan generation, (b) LLM-based action policy, and (c) our LLM-guided search pipeline, which first interacts with the environment for exploration (upper part), then generates a plan based on searching in the state and action space (lower part).}
    \label{fig:overview}
    \vspace{-5mm}
\end{figure*}

From a different perspective, search-based planning is a well-established approach to task planning. Classical algorithms such as breadth-first search (BFS) are capable of finding a path from the initial state to a goal state, where each step corresponds to an atomic action that the agent should take. In essence, a search-based agent is performing trial-and-error within its mental simulation space. It can explicitly reason about the sequence of actions and their consequences, thus leading to a more transparent and sensible decision-making process than the black-box inference of LLMs. Yet, applying search in embodied environments still comes with several challenges: how to obtain a complete and reliable representation of the environment (under partial observability); how to implement accurate state transitions; and how to determine whether a goal state has been reached.

In this work, we aim at utilizing LLMs to implement several modules that assist the search process, addressing the aforementioned challenges while improving search efficiency. A high-level pipeline is presented in Figure~\ref{fig:overview}. Specifically, we propose: (1) a \textbf{T}ranslation module that converts the language instructions into a set of grounded predicates, enabling convenient termination condition checking and task progress monitoring; (2) an \textbf{E}xploration module that leverages the commonsense knowledge of LLMs to guide targeted exploration of the goal-relevant objects; (3) a \textbf{S}imulation module that constructs a rule-based world model with a set of Python functions, allowing for prediction of the state transitions resulting from actions; and (4) a filtering module that progressively prunes the action space to improve \textbf{S}earch efficiency.
The resulting agent, referred to as TESS, effectively combines the performance advantages of search-based planning with the rich prior knowledge encoded in LLMs.
Moreover, by breaking down the planning problem into a series of subproblems, TESS enables a modular decision process that is more interpretable and transparent, allowing us to take a closer look at the effectiveness and limitations of LLMs in different stages of embodied planning.

In brief, the main contributions of this work include: (1) We propose TESS, an LLM-guided search-based pipeline for Embodied Task Planning. (2) We build a new dataset based on the VirtualHome Simulator. It encompasses tasks of varying difficulty, human-like instructions, and diverse environments, serving as a suitable benchmark for automated and accurate evaluation of planning. (3) Extensive experiments on the proposed benchmark and ALFWorld validate the effectiveness of TESS.

%% file: sec/2_related.tex
\vspace{-6mm}
\section{Related Works}
\label{sec:related}
\vspace{-2mm}
\subsection{Embodied Task Planning}  
Automated planning, both in general domains~\cite{STRIPS,PlanGraph} and robotics~\cite{CTAMP,HTAMP}, has a well-established research history. Recently, with the remarkable progress of LLMs across a wide range of tasks, LLM-based approaches have emerged as a prominent paradigm for embodied task planning. Pioneering works prompt off-the-shelf LLMs to translate the instruction into a complete plan~\cite{LMZSP}, or finetune LLMs with expert trajectories and action experiences collected from the environment~\cite{LMIDM}. Upon them, subsequent research has explored several directions: (1) designing new data construction strategies and training objectives, such as self-supervised learning~\cite{LMWM,BPO}, tool-using paradigm~\cite{ToolkenGPT,Toolken+,AgentGym}, knowledge distillation~\cite{plasma}, reinforcement learning~\cite{LLMGP,UETPA} and novel versions of preference optimization~\cite{SPO,TnE,IPR,D2PO}; (2) addressing the limitations of next-token prediction by introducing test-time scaling mechanisms, such as decoding guided by a reward model or discriminator~\cite{SayCanPay,plasma,DGEP,ARMAP,LAC}, and inference based on tree search~\cite{LLMCK,TOOLCHAIN}; (3) leveraging environment feedbacks for reflection and self-improvement, where LLMs receive signals from the simulator's error messages~\cite{CRP,CAPE,CSF} or self-evaluation modules~\cite{E2CL,ESM}, then perform local or global replanning accordingly; (4) introducing diverse representations for the generated plan, such as codes~\cite{PROGPROMPT,FLTRNN,Parsel}, tree structures~\cite{TreePlanner,IIUOBP}, procedures~\cite{NeSyPr}, or hierarchical skills~\cite{SSG,MLDT}, that can encourage agents to actively acquire information during execution and to recover from failures more effectively; and (5) designing more informative and targeted prompts to enhance in-context learning~\cite{NSPP,GEAP,LoTa-Bench}, along with applying post-processing to the actions generated by the LLM.

Instead of directly generating plans using LLMs, our method leverages search to mitigate several common issues associated with LLM inference, such as the omission of environment constraints and the lack of interpretability in the decision-making process.
Some prior works~\cite{RAP,LLMCK,TOOLCHAIN,TreePlanner,LATS,ToS} also feature a ``search'' process. However, the notion of search in those contexts typically refers to the decoding strategies over the \textit{output space} of LLMs. In contrast, the search we propose is conducted directly in the \textit{state and action space} of the environment. Given access to a reliable world model, our approach can theoretically guarantee the correctness and optimality of the resulting plan.

\subsection{LLM and Classical Planning}
PDDL~\cite{PDDL} provides well-established formalisms for modeling and solving planning problems, with extensions~\cite{PDDL2.1,ProbPDDL,ConFF} supporting richer settings such as temporal and numeric constraints, probabilistic or non-deterministic actions, and partial observations.
The complementary nature of these classical planners and LLM-based generative planning is clear. The former provide strict guaranties on the quality of the results and follow a well-formulated decision procedure, but heavily rely on expert-crafted domain and task specifications. LLMs leverage commonsense knowledge to generate responses in a zero-shot manner, but their decision-making process is often opaque and lacks correctness guarantee. A line of work~\cite{LLMAPS,LLMHP,LLMPF} has explored bridging the two sides by utilizing LLMs to construct symbolic planning specifications, such as PDDL domain files~\cite{LLMPDG,Proc2PDDL,LASP,PlanDark,Text2World,GSWM,LMCIAS}, problem files or goal specifications~\cite{TNLPG,LLM+P,PlanCollabNL,VLIRTP,TwoStep,DPLLM,LLM-Regress,Planetarium,PIP-LLM}, or both~\cite{CUWM,AutoTAMP,ISR-LLM,LAPR,AutoGPT+P,InterPreT,GCPDDL,NL2Plan,LEI,DELTA,LaMMA-P,LLMFP,limitLMPF,STAR}, from language descriptions.
Among them, several approaches~\cite{DPLLM,LASP,LLM-Regress} further leverage LLM reasoning to iteratively diagnose execution failures, infer plausible world states, or predict object affordance.
However, most of these methods are still largely confined to small-scale, classical planning problems (\textit{e.g.}, IPC domains), and struggle to handle scenarios that involve a large number of objects.
We aim to achieve a deeper integration of LLMs and classical methods to enable more effective LLM-assisted planning, without manually specified domain or problem files. 
Rather than using the PDDL as the sole interface between the two components, we incorporate the LLMs into multiple stages of the planning pipeline, thus leveraging their strengths more comprehensively to address more complex tasks.

%% file: sec/3_method.tex
\vspace{-3mm}
\section{Methods}
\vspace{-1mm}
\label{sec:method}
\subsection{Problem Formulation}
\vspace{-1mm}

The problem of embodied task planning places an agent within an interactive environment $\mathcal{E}$. At each time step $t$, the agent receives an observation $o_t$ of the current environment state $s_t$, and executes an atomic action $a_t$ selected from its action space $\mathcal{A}$, thereby altering the environment. An instruction $I$ is provided at the beginning of the episode, and the objective of the agent is to accomplish the goals specified by $I$ through a sequence of actions. 

We adopt a setting of text-based environment, where the observations and actions are in the form of natural language. 
Specifically, we mainly focus on the VirtualHome simulator (VH)~\cite{VirtualHome}. It represents the environment as a scene graph (SG), $\mathcal{G}=(V_{\mathcal{G}},E_{\mathcal{G}})$, where each node $n\in V_{\mathcal{G}}$ is an object with attributes and properties, and each edge $e\in E_{\mathcal{G}}$ is a spatial relationship between two objects. VH implements several types of actions and each type may take $0\sim2$ objects as arguments to form an action instance (\textit{e.g.}, \verb+[StandUp]+, \verb+[Grab] mug+, \verb+[PutBack] mug table+).

\begin{figure*}
    \centering
    \includegraphics[width=0.9\linewidth]{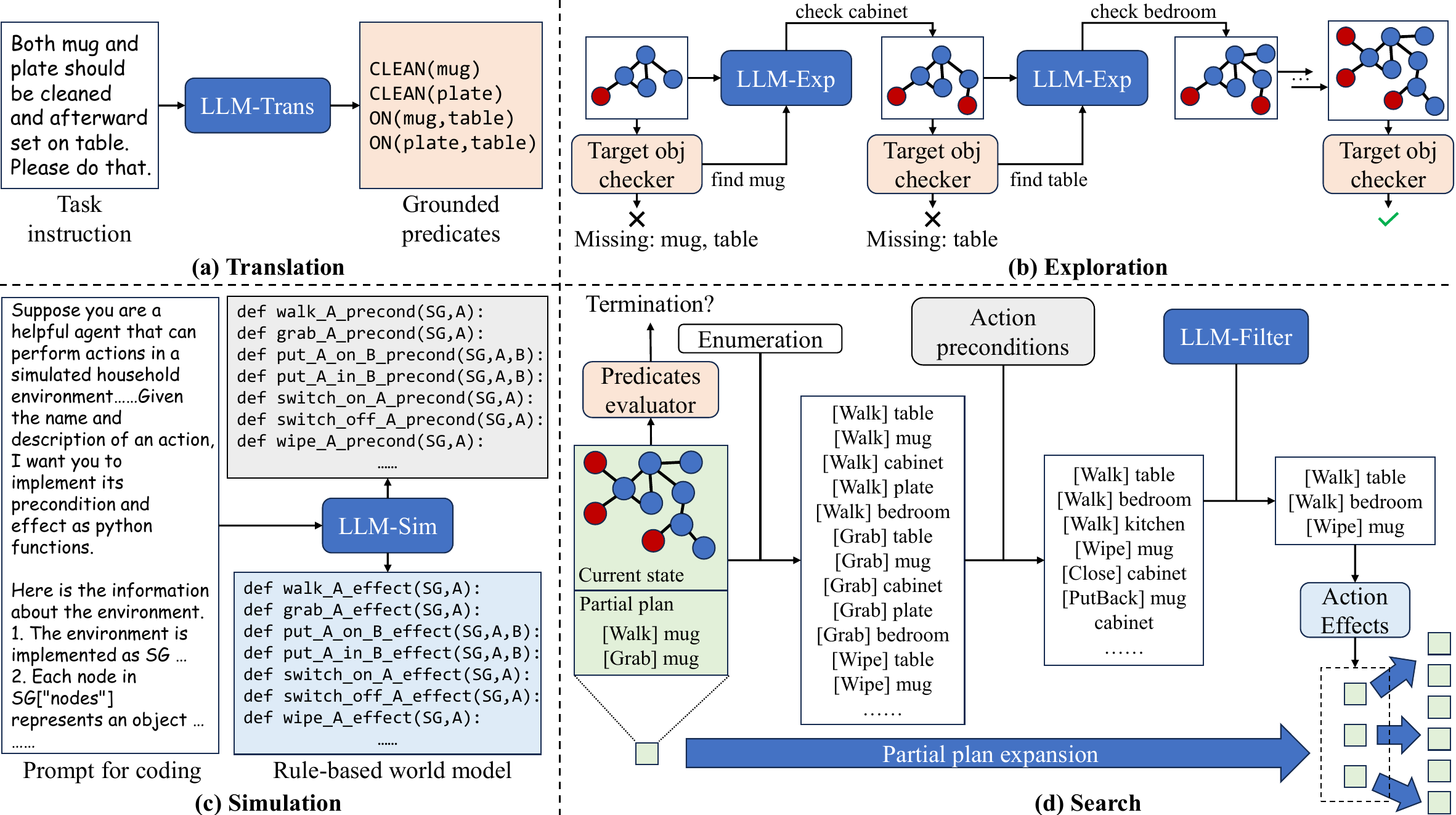}
    \vspace{-2mm}
    \caption{A visualization of the four LLM-based modules in our pipeline.}
    \label{fig:pipeline}
    \vspace{-5mm}
\end{figure*}

\vspace{-1mm}
\subsection{Overall Pipeline}
\vspace{-1mm}
The key idea of this study is to address the planning problem using search algorithms. By defining the search space as the set of all possible environment states, and considering two states to be adjacent if an atomic action can transition one into the other, algorithms like breadth-first search (BFS) are capable of finding a path from the initial state to a goal state. This path can then be translated into a plan, which is guaranteed to be executable, correct, and even optimal. 

However, implementing this search process relies on three essential conditions. (1) The environment must be fully observable; otherwise, the initial state cannot be determined, and the search cannot begin. (2) The state transition resulting from each atomic action must be known; otherwise, the search cannot proceed by extending paths. (3) A mechanism must exist for determining whether a given state is a goal state (\textit{i.e.}, satisfies the instruction requirements); otherwise, the search cannot terminate. These conditions are generally unsatisfiable for complex environments like VH. 

To address these problems, we leverage the commonsense knowledge of LLMs to make the search process feasible and efficient. The overall pipeline is shown in Figure~\ref{fig:overview} (c) and Algorithm~\ref{alg:TESS} (Appendix). Given a task, the translation module translates the language instruction into a set of structured predicates. Then, the exploration module guides the agent to explore the environment until all goal-relevant objects have been observed. Finally, using an LLM-generated simulator and an LLM-based action filter, the search module performs search to generate a plan, which is then executed by the agent within the environment. The modules involved are illustrated in Figure~\ref{fig:pipeline}, and will be detailed in the following subsections.

\vspace{-1mm}
\subsection{LLM-guided Translation and Exploration}
\vspace{-1mm}
\label{sec:method-TE}
The translation module is responsible for the termination mechanism. A naive idea is to query an LLM with the current state and the instruction, and let it decide whether the agent should terminate. 
This approach requires invoking LLM at every step of the search process, which incurs a relatively high computation cost. Moreover, the hallucination of LLMs may lead to unreliable assessments of task completion, especially in complex environments (the input state representation is long) with complex instructions. To enable a more efficient termination criterion, we implement a goal state checker based on a set of grounded predicates. Each one involves either a unary predicate, which checks whether the attributes of an object meet specified conditions, or a binary predicate, which checks whether the relationship between two objects satisfies the given requirements. At the beginning of each episode, the LLM translates the instruction into grounded predicates. During the search process, these predicates are applied to the state. Once all predicates are evaluated as true, the search can be terminated. This translation-based approach decouples termination detection from the search procedure, significantly reducing the computational overhead of this step. Moreover, by decomposing the instruction into a set of subgoals, it enables clearer monitoring of task progress throughout the search process.

Next, to solve the problem of partial observation, the agent needs to actively explore the environment and aggregate its sequential partial observations to build a full representation of the environment. Formally, the agent maintains an SG, $\mathcal{G}^{\text{agent}}$, in its memory. $\mathcal{G}^{\text{agent}}$ is initialized as the first observation $o_0$ at the beginning of each episode and is continuously expanded and updated with the following observations.
We note that constructing a complete SG is actually not necessary, as many objects in the environment may be irrelevant to the task and do not need to be considered during execution. A useful byproduct of the instruction-to-predicate translation process is a set of task-relevant target objects $O^{\text{target}}\subseteq V_{\mathcal{G}}$ (\textit{e.g.}, all the arguments appearing in the predicates). We can safely stop the exploration at timestep $t$ if all target objects have been observed in $\mathcal{G}^{\text{agent}}_t$, which can then serve as a valid and sufficient initial state for the search algorithm.

The key consideration of the exploration stage lies in the interaction efficiency. A random exploration policy may waste many steps on opening irrelevant containers or moving between rooms without meaningful progress. To accelerate this process, we utilize the LLM to generate more human-like exploration trajectories. Specifically, at time step $t$, if some target object $o\in O^{\text{target}}$ is not yet observed in $\mathcal{G}^{\text{agent}}_t$, we provide the LLM with the name of $o$ and a list of candidate objects from $\mathcal{G}^{\text{agent}}_t$ that are still unexplored and potentially contain other objects. The LLM is then prompted to select one object from the list as the next exploration direction. By exploiting the LLM’s prior knowledge of typical room layouts and object placements, our approach enables efficient, goal-directed exploration without the need for exhaustive traversal of the entire environment.

\vspace{-1mm}
\subsection{LLM-guided Simulation}
\vspace{-1mm}
\label{sec:method-Sim}
To simulate state transitions, we need a world model $\mathcal{W}$ for predicting the outcome of an atomic action: $s_{t+1}=\mathcal{W}(s_t, a_t)$. 
A straightforward idea is to implement $\mathcal{W}$ by calling an LLM. However, this introduces the same computational cost and hallucination issues as querying the LLM for termination decisions.
Instead, we aim to fully exploit the structure of the agent's action space to develop a more compact world model. It could be noted that the complexity of the action space primarily arises from the diversity of objects, while the type of actions an agent can perform is relatively limited. For example, VH implements a few dozen of action types like walk, open, sit, and so on. Actions within the same category typically result in similar outcomes (\textit{e.g.}, open a fridge and open a cabinet). Based on this, we leverage an LLM to generate a Python function for each action type. Each function takes as input the current environment state $\mathcal{G}$ and the objects in the action's argument, modifies the nodes and edges of $\mathcal{G}$ according to the expected effects of the action, and returns the updated $\mathcal{G}$ as a prediction of the next state. To guide the LLM in generating reliable functions, we carefully design a prompt that includes a task description, the input-output format, a detailed explanation on the structure of $\mathcal{G}$, and a manually written example function (please refer to the Appendix for the whole prompt). The resulting set of functions forms a rule-based world model, enabling us to simulate state transitions more efficiently and with greater interpretability.
In addition to the effect functions, we also implement a precondition function for each action type using a similar approach. The precondition function also takes state $\mathcal{G}$ and the action's argument as input, and returns a Boolean value indicating whether the action is executable under the given state. These functions will be integrated into the search process described in the following subsection. 

Modeling actions through preconditions and effects is a widely adopted practice in classical planning, such as in the PDDL framework. 
Without generating complete PDDL specifications, our method efficiently implements state transitions through Python functions and scene graph representations.
Moreover, it provides greater flexibility for integration with various search algorithms, particularly those guided by LLMs.

\begin{table*}
  \centering
  \scalebox{0.85}{
  \begin{tabular}{llccccccc}
  \toprule
   & & Obs & \multicolumn{2}{c}{Seen}  &  \multicolumn{2}{c}{Unseen\_obj} &  \multicolumn{2}{c}{Unseen\_task}   \\ 
    \cmidrule(r){4-9}& &  & SR & GCR & SR & GCR & SR & GCR \\ 
    \midrule
  ZS & CoT\cite{CoT}+ZSP\cite{LMZSP} & full & 75.5 & 86.3 & 46.0 & 66.7 & 73.4 & 82.5 \\ 
   & ReAct\cite{ReAct} & partial & 55.2 & 69.5 & 21.7 & 49.1 & 53.7 & 66.0 \\ 
   & Reflexion\cite{Reflexion} & partial &  59.0 & 68.5 & 36.0 & 53.6 & 64.7 & 71.8 \\ 
    & ProgPrompt\cite{PROGPROMPT} & obj & 49.9 & 61.4 & 24.2 & 43.6 & 50.9 & 59.8 \\ 
    & FLTRNN\cite{FLTRNN} & full & 71.9 & 83.2 & 47.8 & 66.2 & \textbf{75.0} & 84.1 \\ 
    & TP (upper bound)\cite{TreePlanner} & obj & 64.5 & 79.3 & 37.9 & 67.5 & 70.3 & 83.2 \\ 
    & TESS-GPT (Ours) & partial & \textbf{85.1} & \textbf{90.9} & \textbf{71.4} & \textbf{83.1} & 74.7 & \textbf{86.4} \\    
    \midrule
FT 
& LID~\cite{LMIDM} & partial & 74.2 & 82.4 & 48.4 & 61.6 & 53.4 & 62.9 \\ 
& Equilibrium~\cite{ESM} & obj & 69.1 & 81.8 & 45.3 & 71.0 & 59.4 & 74.3\\ 
& TESS-Qwen (Ours) & partial & \textbf{87.2} & \textbf{92.4} & \textbf{72.0} & \textbf{81.9} & \textbf{70.3} & \textbf{79.4} \\  
\bottomrule
  \end{tabular}
  }
  \vspace{-2mm}
  \caption{Agent performance on the test splits of the proposed VirtualHome benchmark. The metrics are success rate (SR, whether the task is fully completed) and goal conditions recall (GCR, the ratio of completed goal conditions). 
  We classify the compared methods into API-based zero-shot models (ZS) and finetuned open-source models (FT). 
  We also present the type of environment information required by each method, including full (the complete SG as input), partial (partial observation at each step as input), and obj (a list of all objects in the scene). }\label{tab:main} 
  \vspace{-5mm}
\end{table*}

\vspace{-1mm}
\subsection{LLM-guided Search}
\vspace{-1mm}
\label{sec:method-Search}
Given the termination condition, the initial state, and the action transitions, the search algorithm is invoked to generate a complete plan. A classical BFS search maintains a frontier $\mathcal{F}$, where each element $x\in\mathcal{F}$ consists of an environment state and a partial plan. At each search step, an element $x=[\mathcal{G}, p]$ with the shortest partial plan $p$ is popped from $\mathcal{F}$, and the algorithm enumerates all possible actions on $\mathcal{G}$. For each action $a$, the corresponding effect function is used to simulate the resulting state $\mathcal{G}'$. If $\mathcal{G}'$ has not been visited before, a new search element $x'=[\mathcal{G}', p+[a,]]$ will be added to $\mathcal{F}$. 
Clearly, when the environment contains many objects, the number of possible actions at each step becomes large, leading to an intractably huge search space. To address the challenge of search efficiency, we adopt the following strategies.

\textbf{First}, we replace BFS with A* search by introducing a heuristic ``distance to goal''. Since we have decomposed the task into a set of grounded predicates, the number of predicates that are currently satisfied serves well as a proxy for task progress. Thus, the number of unsatisfied predicates provides a natural heuristic estimate for the remaining distance. Formally, we augment the elements in $\mathcal{F}$ from $[\mathcal{G},p]$ to $[\mathcal{G},p,f]$, where $f=\text{len}(p)+u*k$, $u$ is the number of unsatisfied predicates in $\mathcal{G}$, $k$ is a hyperparameter. $\mathcal{F}$ is implemented as a priority queue and its key is $f$. We set $k$ to 3 in practice, which is a rough estimate of the number of steps typically required to satisfy a single predicate. 
\textbf{Second}, as discussed in Section~\ref{sec:method-TE}, we can extract a set of task-relevant objects $O^{\text{target}}$ based on the translated predicates. Manipulating objects outside this set does not contribute to the completion of the task. Therefore, when enumerating possible actions, we restrict the choice of arguments within this set, forming a subset of target object-related actions $\mathcal{A}_{\text{obj}}\subseteq\mathcal{A}$. 
\textbf{Third}, with the generated precondition functions, we can easily determine whether each action in $\mathcal{A}_{\text{obj}}$ can be performed in the current environment, and filter out those that are physically infeasible to form $\mathcal{A}_{\text{precond}}\subseteq\mathcal{A}_{\text{obj}}$. 
\textbf{Finally}, we send $\mathcal{A}_{\text{precond}}$ to the LLM and ask it to determine a subset $\mathcal{A}_{\text{LLM}}\subseteq\mathcal{A}_{\text{precond}}$ by filtering out the semantically unreasonable actions for the current context (\textit{e.g.}, close something right after opening it, drop something to a non-target location). We only consider the action choices in $\mathcal{A}_{\text{LLM}}$ as possible next steps, and update $\mathcal{F}$ with them. By sequentially leveraging goal information from the predicates, physical constraints encoded in the world model, and commonsense knowledge embedded in the LLM, the agent’s action space can be significantly reduced, thereby making the search process tractable. The pseudo-code of the search process is shown in the Appendix.

\begin{table*}
  \centering
  \scalebox{0.85}{
  \begin{tabular}{lcccc}
    \toprule
    & TESS-GPT & Reflexion & TESS-Qwen & LID\\
    \cmidrule(r){2-2} \cmidrule(r){4-4}
   & Trans+Exp+Search=Sum &  & Trans+Exp+Search=Sum & \\
    \midrule
\#call & 1+12+51=64 & 30 & 1+20+42=63 & 57 \\
\#token in & 644+4,534+120,137=125,315 & 392,201 & 282+9,350+18,701=28,333 & 200,678 \\
\#token out & 25+61+16,722=16,808 & 4,967 & 18+146+748=912 & 773 \\
    \bottomrule
  \end{tabular}
  }
  \vspace{-2mm}
  \caption{Analysis on the average cost of each module for one task instance.}\label{tab:cost}
  \vspace{-6mm}
\end{table*}

%% file: sec/4_exp.tex
\vspace{-3mm}
\section{Experiments}
\label{sec:experiment}
\vspace{-2mm}
\subsection{A New Benchmark on VirtualHome}
\vspace{-1mm}
\label{sec:data}
VirtualHome~\cite{VirtualHome} (VH) is a popular household simulator for embodied agents. However, existing datasets on VH constructed by previous works either lack task diversity~\cite{WAH,LLMCK,LoTa-Bench} or require human assessment due to the absence of automated evaluation metrics~\cite{LMZSP}. To enable more comprehensive and accurate evaluation, and to provide high-quality training data for finetuning-based methods, we build a new benchmark based on VH's evolving graph simulator~\cite{SEEAAS}. The data generation process proceeds as follows. We begin by manually defining a set of task templates with varying levels of difficulty, such as ``clean\_A", ``put\_A\_on\_B", ``put\_A\_in\_B\_and\_switch\_on\_B", and each template is accompanied with a set of goal conditions. The placeholders in the templates (\textit{e.g.}, ``A", ``B") are then filled with appropriate object categories using the knowledge base provided by VH~\cite{SEEAAS}, resulting in a concrete task instance. For each instance, we randomly select an available scene in VH and randomly initialize the placement of objects within it. Next, we apply a manually designed heuristic policy to complete the task instance in the scene (\textit{i.e.}, satisfy all the goal conditions), ensuring that it is feasible while collecting an expert trajectory. Finally, we use GPT to convert the task name into a human-like language instruction. The proposed benchmark features a diversity of tasks (25 categories), environments (50 apartments), target objects (300+ categories), and instructions. In addition, our data generation pipeline is highly extensible: new types of task can be easily introduced by defining the template and goal conditions, and each type of task can yield an infinite number of instances. Examples and statistics of our data are provided in the Appendix.

When generating task environments, the 50 apartment layouts in VH are randomly split into 30/10/10 for training/validation/test. We define a set of unseen object categories and ensure that the task instances in the training set do not involve any of them. We also define a set of unseen task types for evaluating the agent’s generalizability on unfamiliar tasks. The results reported in the following subsections are based on a test set of 1k tasks and a training set of 3k tasks. Nevertheless, our benchmark can produce significantly larger-scale data. We plan to release a dataset containing about 100k instances to enable broader training and evaluation in future research.

\begin{table*}
  \begin{minipage}{0.33\textwidth}
\centering
\scalebox{0.8}{
\begin{tabular}{lccc}
\toprule
   & Acc & SR & GCR  \\
    \midrule
TESS-GPT & 96.0 & 80.0 & 88.4 \\
w/ GT Trans & 100.0 & 82.0 & 89.7 \\
TESS-Qwen & 89.1 & 80.2 & 87.2 \\
w/ GT Trans & 100.0 & 84.8 & 90.8 \\
\bottomrule
  \end{tabular}
}
  \caption{Performance of the translation module.}\label{tab:trans}
\end{minipage}
\hfill
\begin{minipage}{0.29\textwidth}
\centering
\scalebox{0.8}{
\begin{tabular}{lc}
\toprule
   & \#steps  \\
    \midrule
GT (full obs) & 2.3 \\
TESS-GPT & 19.9 \\
TESS-Qwen & 30.2 \\
random & 66.6 \\
\bottomrule
  \end{tabular}
}
  \caption{Efficiency of the exploration module.}\label{tab:exp}
\end{minipage}
\hfill
\begin{minipage}{0.33\textwidth}
\centering
    \scalebox{0.8}{
  \begin{tabular}{lccc}
  \toprule
   & SR & GCR & \#search  \\
    \midrule
BFS & 64.4 & 79.2 & 180 \\
A* & 78.5 & 88.3 & 112 \\
A*+LLM & 80.0 & 88.4 & 55 \\
\bottomrule
  \end{tabular}
  }
  \caption{Ablation study on the design of the search module.}\label{tab:search}
  \end{minipage}
  \vspace{-5mm}
\end{table*}

\vspace{-2mm}

\subsection{Implementation Details}
\vspace{-1mm}
We consider using OpenAI's GPT API (TESS-GPT) or finetuning Qwen3-8B (TESS-Qwen) for implementing the four LLM-based modules. In the former setting, we employ GPT-4.1 series to strike a balance between efficiency and performance. In the latter setting, we apply LoRA~\cite{lora} to finetune three Qwen3-8B models respectively for translation, exploration, and filtering. For simulation, we retain the GPT-based version, as this stage lacks suitable training data and presents challenges for smaller LLMs. 

We consider a series of zero-shot (ZS) and finetuning-based (FT) methods as baselines. Some approaches require access to a full SG as environment information, and some rely on a full list of object names present in the scene. In contrast, our method is built upon a more realistic setting of partial observations. 
For a fair comparison, all the ZS baselines use the same API and all the FT baselines are finetuned upon an 8B base model.
More details on the prompts, training data and supervision, and designs of the baselines are provided in the Appendix.

\vspace{-2mm}

\subsection{Main Results}
\vspace{-1mm}
As shown in Table~\ref{tab:main}, under both ZS and FT settings, TESS achieves a significant improvement over methods that utilize the same type of observations. Moreover, across most evaluation metrics, it outperforms approaches that have access to full SG~\cite{LMZSP,FLTRNN} and those allow re-planning based on the error messages of the simulator~\cite{Reflexion,TreePlanner}. These results highlight the superiority of our search-based framework compared to directly generating actions with LLMs.

Among different test splits, we observe that prior methods often perform poorly on the Unseen\_obj split. This is somewhat counterintuitive for the ZS setting, as it does not involve training and the performance gap between seen and unseen scenarios is supposed to be small. In fact, as described above, the task instances in Unseen\_obj are the ones that contain target objects from the unseen object list. More challenging tasks (those involving a larger number of target objects) are more likely to include unseen items, which leads to a higher average complexity in the Unseen\_obj split compared to the Seen split. Our method demonstrates a strong performance advantage over baselines on Unseen\_obj, suggesting its effectiveness in handling more difficult tasks.

It is also interesting to compare between the ZS setting and the FT setting. By learning from expert trajectories, FT methods often achieve better performance than their ZS counterparts on the Seen split (\textit{e.g.}, LID vs. ReAct, TESS-Qwen vs. TESS-GPT), though they are based on much smaller open-source models. However, previous FT baselines typically suffer a significant performance drop on the Unseen\_task split, due to their limited ability to generalize to novel task types and unfamiliar action patterns. In contrast, our method maintains a clear advantage on these tasks, highlighting the universality of the classical search-based approaches across different problem domains.

\vspace{-2mm}
\subsection{Ablation Study}
\vspace{-1mm}

We present the average cost of calling LLMs in Table~\ref{tab:cost}. The primary source of LLM calling in TESS arises from the exploration and search modules, as these involve repeated queries to determine the exploration targets and filter the action space. 
In terms of the total number of tokens, our approach demonstrates lower cost than previous LLM policies, such as Reflexion and LID. 
Note that though TESS has more calling times, the cost for each call is lower, since the LLMs are used for modular subtasks that are easier than directly determining the action.
A comparison between GPT and Qwen reveals that the former requires significantly more tokens per call during the translation and search stages, which is a result of including demonstrations and CoT prompts for enhanced in-context learning. The cost for simulation is neglected in the table, which is about only 2k/4k input/output tokens for the whole experiment.

We now turn to a fine-grained examination on the performance of specific modules. For \textbf{translation}, TESS-GPT achieves near-perfect accuracy (Acc) on grounded predicates prediction, while TESS-Qwen attains an accuracy of 89.1\% (Table~\ref{tab:trans}). The drop is primarily due to the generalization on unseen tasks.  When replacing the LLM-guided translation with ground-truth predicates, the overall performance improves marginally, suggesting that the translation module performs reliably within our pipeline.
For \textbf{exploration}, we present the number of steps that the agent spends on exploration in Table~\ref{tab:exp}. TESS with GPT or Qwen completes the exploration with less than half the number of steps required by random policy, highlighting the benefit of leveraging the LLM's prior knowledge about the scene. We also note that GPT results in a more effective explorer than Qwen, which is consistent with the cost of exploration in Table~\ref{tab:cost}. The number of exploration steps under full observation serves as a loose lower bound, as even humans may find it difficult to predict the exact locations of the targets based on partial observations in an environment with diverse items. For the \textbf{search} process, we provide an ablation experiment on the action space pruning techniques in Table~\ref{tab:search}. The maximum number of search steps for each episode is set to 500. Within this budget, replacing BFS with A* search and introducing LLM-filter both contribute to higher performance and less average search steps (\#search).

\begin{table}[]
\centering
  \scalebox{0.85}{
  \begin{tabular}{lcc}
  \toprule
   & Seen & Unseen     \\
    \midrule
   ReAct~\cite{ReAct} & 82.8 & 80.6    \\
    Reflexion~\cite{Reflexion} & 78.6 & 72.4      \\
    AdaPlanner~\cite{AdaPlanner} & 93.6 & 91.8 \\
    Expel~\cite{Expel} & 74.3 & 79.8 \\
    RAP~\cite{RAP_AW} & 87.1 & 90.3 \\
    \midrule
 TESS-GPT (Ours) & \textbf{96.4} & \textbf{98.5} \\
 \bottomrule
  \end{tabular}
  }
  \vspace{-3mm}
  \caption{Agent performance (success rate) on the seen and unseen splits of ALFWorld.}\label{tab:AW}
  \vspace{-6mm}
\end{table}

\vspace{-2mm}
\subsection{Experiment on ALFWorld}
\vspace{-1mm}
ALFWorld~\cite{AW} (AW) is another text-based embodied simulator for task planning. It differs from VH in several aspects. 
For example, AW uses room-level scenes, which are smaller in scale than the apartment-level scenes in VH; however, many tasks in AW involve the use of tools (\textit{e.g.}, put a \emph{cooled} object somewhere), placing higher demands on the translation and exploration modules.
In Table~\ref{tab:AW}, we present the results of our GPT-based agent and several recent training-free methods. All the compared baselines are re-implemented with the same GPT API.
The advantage of TESS in success rate indicates its potential to generalize across different simulators.
Details about the agent in AW are shown in the Appendix.

%% file: sec/5_conclusion.tex
\vspace{-2mm}
\section{Conclusions} 
\label{sec:conclusion}
\vspace{-2mm}
We propose TESS, a search-based method for embodied task planning, which leverages LLMs for translating goal conditions, exploring target objects, simulating state changes, and accelerating the search process. It combines the strengths of classical solvers and LLMs, resulting in a more effective and interpretable decision-making process.

%% file: sec/X_supp.tex
\clearpage

\section{Details of the Environment}
\subsection{VirtualHome Simulator}
\label{sec:supp-VH_sim}
VirtualHome (VH) is an environment originally proposed for analyzing human activities in households~\cite{VirtualHome,SEEAAS,WAH,VHAG,EEPL}. Thanks to its definition of tasks and atomic actions, it has recently become a popular testbed for embodied task planning.
VH provides two types of simulators: one based on Unity and the other on evolving graphs. The Unity-based simulator requires interactions with an external Unity process, enabling physical modeling of actions and rendering of visual observations by placing cameras on the agent or in the environment. The evolving-graph-based simulator runs entirely in Python, representing the environment as a scene graph (SG) and implementing actions through updates to object attributes and inter-object relations. In this work, we adopt the latter simulator, as it supports a broader set of action types (some of them are not yet implemented in the Unity version), and allows for easier automatic generation of task environments. Extending our approach to vision-based planning, which would require the Unity-based simulator, is left for future work.

VH is distributed under the MIT License\footnote{https://github.com/xavierpuigf/virtualhome}. 
In this work, we use version 2.3 of VH, which is the latest available release at the time of submission. An introduction about the simulator is provided in its official document. Here we include several key pieces of information, as they are essential for understanding the environment structure and will be used in the subsequent section of data construction process.
\paragraph{Environment} The whole environment is represented as an SG, which has a set of nodes and a set of edges. Each node represents an object and is implemented as a dictionary. Similarly, each edge encodes a spatial relation between two objects and is also represented as a dictionary.
\paragraph{Observation} VH implements the agent's partial observation as a subgraph of the full SG. Specifically, the agent is only able to perceive objects that are located in the same room and are not within any closed container.
\paragraph{Apartment} VH provides 50 distinct apartments, each with its own room layout and initial object arrangement. The differences between apartments are evident in the Unity-based simulator that includes the vision modality. However, these differences are less prominent in the graph-based simulator, as much of the scene-specific, fine-grained information is abstracted away when the physical environment is represented as an SG. Nevertheless, we take apartment diversity into account when constructing our dataset, in order to prepare for future research on vision-based planning.
\paragraph{Object} Each object in the scene, \textit{i.e.} each node in the SG, contains the following information: class\_name, ID, a list of properties, and a list of states. The ID is an integer uniquely assigned to each object instance within the scene. VH defines 23 kinds of object properties and 10 kinds of object states. The former (e.g., \verb+grabbable+, \verb+has_plug+, \verb+recipient+) describe inherent properties of the object, while the latter (e.g., \verb+closed+, \verb+sitting+, \verb+clean+) capture the objects' current conditions. In particular, one special node represents the agent, which is always assigned an ID of 1.
\paragraph{Relation} Each relation in the scene, \textit{i.e.} each edge in the SG, contains the following information: from\_id, to\_id, and relation\_type. The from\_id and to\_id are the IDs of the objects that this edge links together. VH defines 7 relation types, \textit{e.g.}, \verb+inside+, \verb+on+, \verb+close+.
\paragraph{Action} VH implements about 40 types of actions. Each action type may take 0-2 object IDs as argument to form an action instance (\textit{e.g.} \verb+[StandUp]+, \verb+[walk] <mug> (123)+, \verb+[PutIn] <mug> (123) <cabinet> (234)+). An action instance can modify the environment's SG by changing the state of certain objects or altering the relations between objects. However, many actions do not lead to any change in the SG, as their semantics cannot be represented by the set of states and relations defined by VH.

\subsection{Data Generation Pipeline}
\label{sec:supp-dataGen}

The general pipeline for generating our dataset is provided in the Experiment section of the paper. In the following, we will give a more detailed description. Some examples are shown in Table~\ref{tab:supp-sample}.
\paragraph{Step 1. Define task templates} To enable the automated generation of a large number of tasks, we first define a set of task templates, each representing a category of tasks. These templates contain placeholders that can be substituted with one or more objects from the scene to produce concrete task instances. For each template, we specify a corresponding set of goal conditions. Each goal condition can either require that some object has some state, or specify a desired relation between two objects.

\paragraph{Step 2. Generate task instances} Given a task template, we make use of the knowledge base provided by VH which specifies the possible states and placing rules for different object classes. Combining these rules with the goal conditions, we can automatically fill the placeholders with object classes. For example, for the template ``clean A'', we obtain a list of objects that may have the state \verb+clean+, and use each of them to implement a task instance. For template ``put A on B'', we obtain all the reasonable \verb+on+ relations from the placing rules, and use each of them to implement a task instance.

\paragraph{Step 3. Initialize environments and ground task instances} Given a task instance that has replaced the ``A/B/C'' with object classes, we need to build an environment and ground the classes to object instances. We begin by initializing one of VH's apartments. Some random objects are then inserted into it according to the placing rules, in order to make the scene more complex. Next, we verify whether all target objects involved in the task instance are already present in the scene. If any are missing, we add them accordingly. We also slightly adjust the states and positions of the target objects to ensure that the initial scene does not already satisfy the goal conditions, which prevents trivial task completion without any actions being taken. The resulting SG will serve as the initial environment state for the task. Based on this SG, we identify the IDs of the target objects and use them to fill the placeholders in the goal conditions. The finalized goal conditions can then be used to evaluate the generated plans.

\paragraph{Step 4. Generate expert trajectories} We design a heuristic policy that solves the goal conditions one by one (in a proper order), and use it to generate expert trajectories. The collected plan may not be the optimal one (\textit{e.g.}, the agent transfers one object at a time under this policy, though it may grab two objects simultaneously), but it is sufficient to satisfy all the goal conditions. This process also serves to verify that the previously generated task instances are reasonable and achievable.

\paragraph{Step 5. Generate instructions} We leverage GPT-4.1 to convert the task instances into diverse and human-like user instructions. In practice, due to the large number of task instances, we actually send the task templates instead of the instances to GPT. Each template will be transformed into various instructions containing placeholders. These placeholders are then replaced with the specific objects from individual task instances. 
When invoking GPT, the goal conditions of the task template are also included in the prompt to prevent the model from generating overly vague instructions or deviating from the intended task objectives.

\paragraph{Step 6. Data split} We first divide the 50 apartments provided by VH into a 30/10/10 split. Based on that, all generated samples are assigned to the training/validation/test sets. Subsequently, we define a subset of task templates as unseen task types, and a subset of object categories as unseen object types. In the training set, we exclude all samples that involve either unseen tasks or unseen objects. For the validation/test set, we construct three subsets: one containing only seen tasks and seen objects, one involving unseen tasks, and one involving unseen objects.

\begin{table*}[h!]
  \centering
  \scalebox{0.8}{
  \begin{tabular}{lll}
  \toprule
  Task template & Goal conditions & Instruction for an example task instance \\
  \midrule
  clean A & A is CLEAN & ``Can you take care of cleaning waterglass-269 for me?'' \\
  turn on A and hold & A is SWITCHON; & ``Please turn on lighter-451 and keep holding \\
  & agent is holding A &  it in your hand.'' \\
  put A on B on C & A is on B; &  ``Put plate-363 on table-575, then arrange food\_fish-594 \\
                & B is on C &  and food\_vegetable-563 to sit on top of plate-363.'' \\
  \bottomrule
  \end{tabular}
  }
  \caption{Examples of the proposed dataset. We first manually define a set of task templates and their corresponding goal conditions. Then we fill the placeholders with proper object categories to get task instances, and use GPT to generate human-like instructions for the tasks. Note that placeholder ``A'' may be replaced by more than one objects to enhance task complexity.} 
  \label{tab:supp-sample}
\end{table*}

\subsection{Dataset Statistics}

The pipeline described above has the potential to generate infinite tasks. In practice, we define 25 templates in Step 1. By enumerating all possible object classes according to the knowledge base and grounding them into different apartments, we obtain 165k grounded task instances in Step 3. After the split in Step 6, we have a training set of size 76k, and the val\_seen, val\_unseen\_obj, val\_unseen\_task, test\_seen, test\_unseen\_obj, test\_unseen\_task splits each has 26k, 11k, 10k, 25k, 10k, 10k task instances.
As stated in the Experiment section of the main text, in order to reduce the computational cost associated with training and evaluation, we sample a mini-scale dataset consisting of 3k training samples, 1k validation samples, and 1k test samples for the experimental results reported in this paper.

Previous works have built various datasets on top of VH. They can be roughly categorized into two types based on their construction methodology: program-based and condition-based. The former~\cite{LMZSP,Parsel,GEAP,ToolkenGPT,ToolBench,SayCanPay,TreePlanner,ESM,E2CL} leverages the ActivityProgram dataset~\cite{VirtualHome} provided by VH, which contains approximately 300 tasks and over 1,000 plans. Each plan consists of a sequence of actions (representing the expert trajectories for task planning), and can be grounded into different apartments in VH. Program-based approach takes full advantage of the rich set of action types in VH and covers a wide range of tasks. However, as noted in Section~\ref{sec:supp-VH_sim}, many actions in VH do not result in any change to object states or relations, making it difficult to determine task completion by the SG. This presents challenges for the automated evaluation of plans.

In contrast, our dataset is constructed using the condition-based approach, where tasks are defined by goal conditions. The agent is required to generate plans that transform the initial SG into a final SG that satisfies the conditions. While this strategy sacrifices some diversity in actions and task types, it allows for a more reliable and accurate evaluation of the model’s planning capabilities. We follow the common practice to define the goal condition of a task as a set of atomic conditions, each focusing on a specific state of an object or a specific relation between two objects. Differently, a recent benchmark, EAI~\cite{EAI}, defines goals using linear temporal logic (LTL), thereby supporting tasks that involve temporal dependencies and requirements on the action process. However, we observe that the goal specifications in EAI exhibit a certain degree of ambiguity, making it difficult for LLMs to accurately infer goal conditions from the task instructions. This is consistent with the reported performance on the benchmark, where Goal Interpretation is a critical bottleneck. Moreover, the dataset is relatively limited in scale, comprising only 26 task instances grounded in 338 environments. Therefore, in this work, we still define the task goal with a set of atomic conditions, and leave the exploration of more expressive goal representations to future work.

\begin{table*}[h!]
    \centering


\scalebox{0.8}{
        \begin{tabular}{lcccccc}
    \toprule
   &  \# act. types & \# cond. types & \# target obj. types & \# apt. & \# task types \\
    
    \midrule
  ActivityProgram~\cite{VirtualHome} & \textbf{40} & - & 196 & 7 &  - \\ 
  LID~\cite{LMIDM} & 7 & 4 & 15 & 1 & - \\    
  ProgPrompt~\cite{PROGPROMPT} & 13 & - & 35 & 3 & - \\ 
  llm-mcts~\cite{LLMCK} & 6 & 2 & 37 & 2 & - \\ 
  LoTa-Bench~\cite{LoTa-Bench} & $\sim$6 & 3 & 16 & 7 & - \\ 
  Ours & 13 & \textbf{11} & \textbf{307} & \textbf{50} & 25 \\
  \bottomrule
  \end{tabular}
  }
  
\vspace{2mm}


\scalebox{0.75}{
        \begin{tabular}{lcccc}
    \toprule
   & \# obj. per scene & \# target obj. per task & \# cond. per task & human-like instr. \\
    
    \midrule
  ActivityProgram~\cite{VirtualHome} & 290 & 1-\textbf{18} & - & Yes \\ 
  LID~\cite{LMIDM} & 273 & 2-11 & 2-\textbf{10} & No \\    
  ProgPrompt~\cite{PROGPROMPT} & 115 & 1-11 & - & Yes \\ 
  llm-mcts~\cite{LLMCK} & $\sim$300 & 2-6 & 1-3 & No \\ 
  LoTa-Bench~\cite{LoTa-Bench} & 357 & 2-7 & 1-6 & Yes\\ 
  Ours & \textbf{437} & 1-6 & 1-8 & Yes \\
  \bottomrule
  \end{tabular}
}

\vspace{2mm}


\scalebox{0.8}{
        \begin{tabular}{lccc}
    \toprule
   & \# tasks & expert traj. & gen. proc. \\
    
    \midrule
  ActivityProgram~\cite{VirtualHome} & 6k & Yes & No \\ 
  LID~\cite{LMIDM} & 300/5k & Yes & Yes \\    
  ProgPrompt~\cite{PROGPROMPT} & 70 & Yes & No \\
  llm-mcts~\cite{LLMCK} & 2k & Yes & Yes \\ 
  LoTa-Bench~\cite{LoTa-Bench} & 350 & Yes & Yes \\ 
  Ours & 5k/\textbf{165k} & Yes & Yes \\
  \bottomrule
  \end{tabular}
}

 \caption{Statistics of different benchmarks on VirtualHome.} 
  \label{tab:supp-data}
\end{table*}

Table~\ref{tab:supp-data} presents some key statistics of our dataset and compares them with those used in prior works. The proposed dataset has the following features.
(1) \textbf{Task diversity}. We define 25 types of tasks (\# task types) that cover a broader range of apartment layouts (\# apt.), target object categories (\# target obj. types), and goal condition types (\# cond. types) compared to existing datasets. For actions, we start from all action types implemented in VH, removing those that do not lead to any change in object states or relations (\textit{e.g.}, \verb+Watch+, \verb+Touch+). We also eliminate redundant actions with similar functionality (\textit{e.g.}, \verb+Walk+ and \verb+Run+, \verb+Wipe+ and \verb+Wash+) and exclude those unsuitable for robotic agents (\textit{e.g.}, \verb+Lie+, \verb+PutOff+). As a result, we retain a final set of 13 actions that are well-defined and useful for the tasks.
(2) \textbf{Task complexity}. Each of our task instances is grounded in a randomly initialized environment that is more complex than those used in previous datasets (\# obj. per scene). The complexity of individual task instances varies across the dataset (\# target obj. per task, \# cond. per task). While some prior works have a higher number of goal conditions or target objects per task, we note that increasing these numbers in our data generation framework is straightforward—we can simply substitute more target objects into the placeholders during task template instantiation (Step 2 in Section~\ref{sec:supp-dataGen}). More importantly, our task templates are specifically designed to support a wider variety of logically coherent goal condition combinations (e.g., plug in A and switch on A, clean A and place A in B). In contrast, existing datasets often construct complex tasks by merely stacking multiple independent simple tasks (e.g., A on B and C on D and E on F). 
Besides, our task instructions are expressed in natural, daily language generated by GPT, which presents a certain level of challenge for the agent to accurately interpret human intentions.
(3) \textbf{Task scale}. The scale of our data (\# tasks) is an order of magnitude larger than existing datasets. Different samples in the dataset have different task instances or different apartment layouts, offering a highly varied and comprehensive set of environments for evaluation.
In addition, we accompany each sample with an expert trajectory, which can be leveraged for finetuning agents.
(4) \textbf{Extendability}. New types of tasks can be generated by introducing additional task templates and the corresponding goal conditions, a process that requires minimal manual effort. We plan to release the code for automatic task generation (gen. proc.), hoping to support the community in constructing richer and more complex task datasets.

\subsection{The ALFWorld Simulator}
Similar to VH, ALFWorld~\cite{AW} (AW) is also a text-based simulated household environment. It is built on top of a PDDL engine and represents both environment observations and actions in the form of natural language sentences. Consequently, the agent must parse the observations into SGs for the object states and inter-object relations. The official AW benchmark provides two test splits (the seen and unseen validation sets) containing 140 and 134 task instances, respectively. The tasks are grouped into 6 categories (\textit{e.g.} Pick \& Place, Examine in Light) and grounded in 120 different rooms. 11 types of actions are involved for completing the tasks.

\section{More on Related Works}
\label{sec:supp-related}
In this section we provide a thorough review of previous task planning methods on VH. We first classify these works by whether they focus on \emph{grounded} planning, \textit{i.e.}, whether the planning process is conditioned on environment information. \textbf{Ungrounded task planning} can be seemed as a pure natural language processing task, which translates the name of a task into its procedure. The earliest work on this branch is ZSP~\cite{LMZSP}. After generating each step using the LLM, it maps the output to the most similar action in the agent's action space based on Sentence-BERT~\cite{SentenceBERT} embeddings. This helps ensure a certain degree of validity for each generated step. However, the overall executability of the final plan still leaves room for improvement. 
Re-Prompting~\cite{CRP} and CAPE~\cite{CAPE} extend ZSP into a closed-loop framework, where the environment feedback is sent to the LLM to trigger replanning whenever a generated action is not executable.  
PLAN~\cite{NSPP} adopts a neuro-symbolic approach to extract relevant commonsense knowledge from a knowledge base and formalize them into the prompt. 
Parsel~\cite{Parsel} first decomposes the task into function descriptions. Then it implements each function modularly and links them together with a Parsel synthesizer. 
ETP-Bench~\cite{VLMZS} evaluates the performance of existing vision-language models (VLMs) in cross-modal embodied task planning. 
LLM-OBTEA~\cite{IIUOBP} represents the plan as a behavior tree. 
ToolChain~\cite{TOOLCHAIN} employs an inference strategy based on A* search. It designs multiple kinds of cost and heuristic functions to guide the decoding process of the LLM. Unlike TESS which performs search within the space of environment states, the search in ToolChain is not grounded into specific scenes, and is mainly aimed at enhancing the quality of LLM's multi-step outputs.

Apart from the aforementioned zero-shot methods, models can also be finetuned on the task-plan pairs. \cite{LMWM} adds a pretraining stage before finetuning on ZSP's data~\cite{LMZSP}, which includes several proxy tasks to enhance the model's embodied capabilities. It also puts forward a new optimization algorithm based on elastic weight consolidation.
ToolkenGPT~\cite{ToolkenGPT} and Toolken+~\cite{Toolken+} follow an LLM tool-using paradigm and treat different action types as different tools. They introduce some tool tokens into the vocabulary of the LLM, which are trained with expert trajectories.
ToolBench~\cite{ToolBench} utilizes programatically curated data to finetune open-source models, and designs a demonstration retriever for in-context learning. 
Following SayCan~\cite{SayCan}, SayCanPay~\cite{SayCanPay} trains a domain-specific Can model to evaluate the feasibility of candidate actions, and a domain-specific Pay model as an estimator of the action's future reward. The scores produced by these two modules will guide the plan generation process of a frozen LLM (the Say model). 
PLASMA~\cite{plasma} investigates distilling planning abilities from LLMs to smaller language models. It generates a set of goal-plan pairs by GPT and trains a T5 model with that, and it also proposes a verifier-guided step-wise beam search for decoding at inference time.
E$^2$CL~\cite{E2CL} constructs plans that contain feedbacks and corrections based on expert demonstrations. These plans are then used for finetuning the model to develop the ability to reflect and self-refine.

On the other hand, \textbf{grounded task planning} requires an in-depth analysis of the environment and generates plans accordingly. The pioneering work LID~\cite{LMIDM} finetunes a GPT-2 for this task. The model takes the goal conditions, the action history, and the current observation as input. The initial training data consists of expert trajectories. The finetuned agent is used to explore the environment, and additional task-trajectory pairs are collected using a hindsight relabeling technique. Based on LID, VP$^2$~\cite{LMPP} extends the text-based observations to egocentric RGB images and trains a GPT-2 with image tokens. SPO~\cite{SPO} further enables the VLM to evaluate its own reasoning process, in order to collect preference data that can be used to further enhance its capabilities.
TWOSOME~\cite{TwoSome} adopts a reinforcement learning (RL) framework and initializes the action policy with an LLM in order to improve sample efficiency. 
SemGro~\cite{SSG} constructs a hierarchical skill database by iteratively composing atomic actions. It also introduces a tunable VLM as a critic to evaluate the executability of the generated plan and to decompose the unexecutable skills into lower-level sub-skills. MLDT~\cite{MLDT} builds a three-level planning pipeline and finetunes LLMs to decompose a task into subgoals, translate each subgoal into subtasks based on observations, and ultimately convert each subtask into executable actions. EquilibriumPlanner~\cite{ESM} devises a self-refinement process based on fixed-point iteration, and it adopts equilibrium modeling to formulate a supervised training pipeline. Similar to SayCanPay~\cite{SayCanPay}, DGAP~\cite{DGEP} trains a discriminator to assess the quality of LLM-generated actions. It can then be incorporated into the decoding process by adding the score of candidate actions into the prompt or trigger replanning when a predicted action is of low quality. Similar to E$^2$CL~\cite{E2CL}, STeCa~\cite{STeCa} collects calibrated trajectories that include an error correction process, and leverages them to train the agent via RL.

As for zero-shot grounded task planning, \cite{GEAP} augments the input of ZSP~\cite{LMZSP} with environment information, and devises a scoring mechanism to associate the generated object class with a proper object instance.
ProgPrompt~\cite{PROGPROMPT} asks the LLM to generate a complete plan in the form of codes. Atomic actions are executed through function calls, while assert statements are used to verify whether certain conditions are satisfied in the environment. This design enables the planner to proactively acquire and utilize environment information during plan execution. LLM-MCTS~\cite{LLMCK} focuses on the object rearrangement task. It follows the RAP~\cite{RAP} framework and employs the LLM as both the policy and the world model, enabling Monte Carlo Tree Search (MCTS) during action generation to yield better results than simple next-token prediction. Its implementation of the world model involves repeatedly querying the LLM for the potential locations of each object. However, this approach incurs high API costs, requires access to a complete list of all objects and containers within the scene, and struggles with tasks beyond pick-and-place (\textit{e.g.} those involving object state change). In contrast, our rule-based world model simulates actions in a more compact and efficient manner, allowing for convenient handling of a broader range of tasks. FLTRNN~\cite{FLTRNN} first splits the task instruction into subgoals, then solves each of them sequentially with a language-based RNN. TreePlanner~\cite{TreePlanner} focuses on efficient replanning. It generates a batch of plans and merges them into a tree. Then it performs grounded decisions on the tree to find a path that leads to the completion of the task. Note that the search on the tree of LLM-generated actions is essentially different from our search in the space of possible environment states. VoteTreePlanner~\cite{Vote-Tree-Planner} further integrates ProgPrompt’s code-based plans~\cite{PROGPROMPT} with the overall pipeline of TreePlanner. LoTa-Bench~\cite{LoTa-Bench} feeds the instruction and action history into the LLM to compute the conditional probabilities of different candidate actions, and selects the action for the next step based on that. BrainBody-LLM~\cite{CSF} includes a Brain-LLM for parsing the task into subtasks, and a Body-LLM for converting each subtask into atomic actions. It also leverages environment feedbacks for replanning. NL2Plan~\cite{NL2Plan} utilizes LLMs to generate PDDL files given the task description and the environment information, then it calls a classical solver to generate the plan. LookPlanGraph~\cite{LookPlanGraph} employs an VLM to analyze the agent’s visual observations and uses the results to update the SG. SG-RwR~\cite{SG-RwR} proposes an LLM-based retriever that extracts useful information from the SG when needed by the LLM-based planner.

\section{Details of The Zero-Shot Agent}
\label{sec:supp-ZS}
\subsection{Translation}

Figure~\ref{lst:trans} is the prompt for LLM-based translation. It contains a description on the role of the LLM, the possible predicate types, and some restrictions on the output format. We also provide several in-context learning examples. Upon receiving the output from the LLM, we try to parse it into a list of grounded predicates and perform basic validity checks, such as verifying whether the predicate types come from the predefined set. If any issues occur, the LLM is prompted to regenerate the output. For VH, we define the set of target objects as all objects mentioned in the parsed predicates. For AW, where changes in object states often require the use of additional tools (\textit{e.g.}, a fridge is needed to bring an apple to a \verb+COOL+ state), the target object set also includes the tools associated with the state-changing predicates (\textit{e.g.}, ``something that can cool an apple''). Another difference in AW is that it does not specify object IDs in the instruction. Thus, we assign some temporary IDs to the target objects during the translation process. They will later be replaced with the actual instance IDs during exploration. By default, the translation module is implemented with GPT-4.1-mini.

\begin{figure*}[h!]
    \centering
    \begin{lstlisting}[
        basicstyle=\ttfamily\small, 
        frame=single, 
        backgroundcolor=\color{gray!5}, 
        breaklines=true, 
        captionpos=b, 
        columns=fullflexible 
    ]
Suppose you are a helpful agent that can perform actions in a household environment. Upon receiving a task expressed in natural language, I want you to formulate it into a list of facts that should be accomplished at the end of the task execution. 
There are two types of facts: (1) unary fact that descibes a state that an object should possess; (2) binary fact that describes a relation between two objects.
The possible states include: [OPEN, CLOSED, CLEAN, DIRTY, SWITCHON, SWITCHOFF, PLUGGED_IN, PLUGGED_OUT]
The possible relations include: [on, inside, close_to, holding, sitting_on]
You should output exactly a list of facts without other description or explanation. Each item in the list should be a tuple of the form (state, object-id) for unary fact, or a tuple of the form (relation, object1-id1, object2-id2) for binary fact.
Note that the state and relation should come from the lists I provided above. You should focus on the final state of the environment when extracting facts, instead of the process of the actions.
The state of [OPEN, CLOSED, CLEAN, DIRTY, SWITCHON, SWITCHOFF, PLUGGED_IN, PLUGGED_OUT] can only be applied to objects. The relation of [holding, sitting_on, close_to] can only be between the agent (whose id is "agent-1") and an object. 

Here are some examples.
Task: Could you take plate-245 and plate-143, clean them up, and put them on coffeetable-215 for me?
Facts: [(on, plate-245, coffeetable-215), (on, plate-143, coffeetable-215), (CLEAN, plate-245), (CLEAN, plate-143)]

[more examples]

Now comes the query.
Task: [the task instruction]
Facts: 
    \end{lstlisting}
    \caption{Prompt for TESS-Translation.}
    \label{lst:trans}
\end{figure*}

\subsection{Exploration}

In the exploration stage, we aggregate the partial SG observed at each time step into the agent's memory $\mathcal{G}^{\text{agent}}$. To query the LLM, we construct a candidate set of objects for exploration based on the current memory. Each object in this set satisfies the following criteria: (1) it is either a room or a closed, openable container (\textit{i.e.}, it may contain a target object inside); and (2) it has not been explored in previous time steps. 
Figure~\ref{lst:exp} shows the prompt template used during the exploration. In this stage, the LLM only needs to output a single object. We do not provide any examples in the prompt, relying entirely on the LLM’s commonsense knowledge about object placements. By default, the exploration module is implemented with the lightweight GPT-4o-mini.

\begin{figure*}[h!]
    \centering
    \begin{lstlisting}[
        basicstyle=\ttfamily\small, 
        frame=single, 
        backgroundcolor=\color{gray!5}, 
        breaklines=true, 
        captionpos=b, 
        columns=fullflexible 
    ]
Suppose you are a helpful agent in a household environment. I want you to help me find some object. I will give you a target object and a list of observed rooms and objects. The target object is usually hidden in a closed container or storage furniture. You need to choose a room or object in the given list to explore, which should be some place that the target object is likely to be located in. 
Note that your output should be chosen from the given list. Please output directly the room or object that you choose (in the form of "class_name-id"). Do not output any other information.
Hint: for big objects, you should usually choose a room to explore; while for small objects, you may choose a container that may have it inside.

Target object: [an unobserved target object]
Observations: [a set of exploration candidates]
Output: 
    \end{lstlisting}
    \caption{Prompt for TESS-Exploration}\label{lst:exp}
\end{figure*}

As mentioned in the previous section, AW's target object set contains some undetermined tools. For each exploration step aiming for a tool, we first let the LLM determine whether any of the currently observed objects can serve the required function. If such a tool is already observed, the LLM outputs its name. Otherwise, it selects a direction for further exploration in a similar manner as VH. On the other hand, the scenes in AW usually contain a large number of instances belonging to the same category (\textit{e.g.} more than 10 drawers). If we include all these instances in the input, the LLM is likely to hallucinate and generate a non-existing instance ID (\textit{e.g.}, select drawer-2 from drawer-1, drawer-3, drawer-5, drawer-6, drawer-7). Therefore, for agents operating in AW, we only provide the categories of candidate objects as input. Based on the category predicted by the LLM, we then randomly select a valid and previously unexplored instance ID as the exploration direction.

After obtaining an object or room for exploration, the agent will walk to it and open it if it is closed. $\mathcal{G}^{\text{agent}}$ will then be updated with the new observation.

\subsection{Simulation}

The prompt for LLM-based simulation (Figure~\ref{lst:sim}) consists of a detailed description of the environment SG, which will serve as the input of the generated Python functions. We also include some important rules of the simulator, and a pair of handwritten precondition and effect functions as demonstrations. We provide detailed comments in the demonstrations to encourage the development of a coherent chain of thought in the LLM.
We generate the preconditions and effects for all actions within a single LLM call, in order to maintain consistency across the simulated behaviors of different actions. By default, the simulation module is implemented with GPT-4.1. It is invoked only once for the whole experiment.

\begin{figure*}[h!]
    \centering
    \begin{lstlisting}[
        basicstyle=\ttfamily\small, 
        frame=single, 
        backgroundcolor=\color{gray!5}, 
        breaklines=true, 
        captionpos=b, 
        columns=fullflexible 
    ]
Suppose you are a helpful agent that can perform actions in a simulated household environment. Each atomic action that you can perform should have a precondition and an effect. Given the name and description of an action, I want you to implement its precondition and effect as python functions.
Here is the detailed information about the environment.
1. The environment is implemented as a graph. The graph is a python dictionary, whose keys are "nodes" and "edges". graph["nodes"] and graph["edges"] are both lists of dictionaries. 
2. Each node in graph["nodes"] represents an object or a room in the environment. Its attributes include "id" and "states". node["id"] is a string, e.g., "apple-1", "fridge-123". There is a specal node whose "id" is "agent-1", which represents the agent that performs actions. node["states"] is a list of string, representing the current states or properties of the object or room.
3. All the possible states (i.e., the elements in node["states"]) are: [OPEN, CLOSED, CLEAN, DIRTY, SWITCHON, SWITCHOFF, PLUGGED_IN, PLUGGED_OUT, HOLDABLE, SITTABLE, CAN_OPEN, HAS_PLUG, HAS_SWITCH, SITTING]. Note that they are all in upper case. 
[More descriptions about the edges, the relation types, and the actions.]

We implement the preconditions and effects in the form of python functions.
Specifically, the precondition function of an action should take the current environment graph and the arguments of the action as input. It should return a boolean value indicating whether the precondition is satisfied.
The effect function of an action should also take the current environment graph and the arguments of the action as input. It should return the modified environment graph after performing the action.

Here is an example of the required functions. You should follow this code format.
[a handwritten precondition function and effect function]

Here are the actions you need to solve.
[action names and descriptions]

Hint: 1. You need to get close to A if you want to hold or interact with it.
2. Since you have only two robotic arms, you can hold at most two objects.
    \end{lstlisting}
    \caption{Prompt for TESS-Simulation}\label{lst:sim}
\end{figure*}

\subsection{Search and Filtering}

Before expanding the partial plan at each search step, we prune the action space by leveraging goal information from the predicates, physical constraints encoded in the preconditions, and commonsense knowledge embedded in the LLM. The prompt for the LLM-based filtering process is shown in Figure~\ref{lst:filter}. It includes the role of the LLM, a description of the input and output format, some examples, and some rules of the simulator. The actual query consists of the translated predicates that serve as the goal, the current SG converted into a descriptive sentence, and a set of candidate actions that have been filtered based on the target objects and the preconditions. The output is a further filtered set of actions. By default, the filtering module is implemented with GPT-4.1-mini.

\begin{figure*}[h!]
    \centering
    \begin{lstlisting}[
        basicstyle=\ttfamily\small, 
        frame=single, 
        backgroundcolor=\color{gray!5}, 
        breaklines=true, 
        captionpos=b,
        columns=fullflexible 
    ]
Suppose you are a helpful agent performing household actions in a house. I will give you (1) a task and its goal that you need to achieve with a series of atomic actions, (2) the current state of the environment; (3) a set of candidate actions that you can do currently.
I want you to filter the candidate actions by reserving the actions that may be helpful for completing the task from current state and removing the actions that seem useless.

Here are some examples.
(1) 
Goal: blender-571 is SWITCHON, blow_dryer-583 is SWITCHON, blow_dryer-584 is SWITCHON, blender-571 is on table-581, blow_dryer-583 is on table-581, blow_dryer-584 is on table-581. 
Current state: kitchen-115 is DIRTY. bookshelf-228 is DIRTY. closet-49 is OPEN and CAN_OPEN and DIRTY. blender-571 is PLUGGED_IN and CLOSED and CAN_OPEN and HAS_PLUG and HAS_SWITCH and HOLDABLE and DIRTY and SWITCHON [......] agent-1 is close_to bookshelf-228. agent-1 is close_to blender-571. table-581 is inside kitchen-115. folder-199 is inside kitchen-115. folder-199 is inside bookshelf-228. agent-1 is inside kitchen-115. blender-571 is inside kitchen-115 [......]
Candidate actions: ['walk to table-581', 'walk to livingroom-11', 'open blender-571', 'clean folder-199', 'switch off blender-571', 'plug out blender-571', 'grab blender-571', 'put folder-199 on bookshelf-228', 'put folder-199 on blender-571', 'put folder-199 inside bookshelf-228']
Analysis: The goal is to switch on blender-571, blow_dryer-583, blow_dryer-584, and put them all on table-581. Currently, blender-571, blow_dryer-583, blow_dryer-584 are all switched on, so I just need to move them to the table. I am now holding folder-199, which is unrelated to the goal. I should put it back to somewhere. I am also close to blender-571, so another proper next step should be grabbing blender-571, so that I can take it to table-581.
Filtered actions: put folder-199 on bookshelf-228, put folder-199 on blender-571, put folder-199 inside bookshelf-228, grab blender-571

[more examples]

Hint:
1. After "Filtered actions: ", you should output a sequece of actions separated by ", ". **Please make sure that all the acions are selected from the candidate actions.**
2. All the actions in "Candidate actions" are physically performable. You only need to filter those that are not helpful for the current state.
3. You need to be close to an object (by walking to it) before you interact with it.
4. You can hold at most two objects simultaneously.

Now here comes the query.
Goal: [translated predicates]
Current state: [current simulated SG in the form of natural language]
Candidate actions: [a list of executable actions]
    \end{lstlisting}
    \caption{Prompt for TESS-Filter}\label{lst:filter}
\end{figure*}

\section{Pseudo-Code}
Algorithm~\ref{alg:TESS} presents the overall pipeline of the TESS agent. Algorithm~\ref{alg:search} presents the detailed procedure of the search module.

\begin{algorithm*}[h]
\caption{The planning process of TESS agent}
\label{alg:TESS}
\begin{algorithmic}[1]   
\Require Instruction $I$, environment $\mathbf{E}$

\State Precond\_func, Effect\_func = \textcolor{red}{Simulation}() 

\State $g = \text{\textcolor{purple}{Translation}}(I)$ 
\State $O^{\text{target}} = \text{parse\_target\_objects}(g)$ 

\State $\mathcal{G}^{\text{agent}} = \text{initial\_observation}(\mathbf{E})$
\For{$o \in O^{\text{target}}$}
    \While{not $\text{exists\_in}(o, \mathcal{G}^{\text{agent}})$}
        \State $a = \text{\textcolor{orange}{Exploration}}(\mathcal{G}^{\text{agent}}, o)$
        \State $\mathcal{G}^{\text{obs}} = \text{perform\_action}(\mathbf{E}, a)$
        \State $\mathcal{G}^{\text{agent}} = \text{update\_SG}(\mathcal{G}^{\text{obs}}, \mathcal{G}^{\text{agent}})$
    \EndWhile
\EndFor

\State $p=\text{\textcolor{blue}{Search}(Precond\_func, Effect\_func}, \mathcal{G}^{\text{agent}}, g\text{)} $
\For {$a \in p$}
    \State $\text{perform\_action(}\mathbf{E}, a\text{)}$
\EndFor
\end{algorithmic}
\end{algorithm*}

\begin{algorithm*}[h]
\caption{The \textcolor{blue}{Search} module of TESS}
\label{alg:search}
\begin{algorithmic}[1]  
\Require Precondition functions $P$, effect functions $E$, initial state $\mathcal{G}^{\text{init}}$, goal predicates $g$, action space $\mathcal{A}$, hyper-parameter $k$, search budget $N$

\State $\mathcal{F} = \text{Priority\_Queue}()$
\State $\mathcal{F}\text{.push((}\mathcal{G}^{\text{init}}, \text{[]}, 0\text{))}$
\State $visited$ = Set()\;
\State $n_{\text{best}}=0$
\While{$\mathcal{F} \neq \emptyset$}
    \State $(\mathcal{G}, p, f) = \mathcal{F}\text{.pop()}$
    \If {$\mathcal{G}\in visited$}
        \State \textbf{continue}
    \EndIf
    \State $visited$.add($\mathcal{G}$)
    \State $s = \text{get\_satisfied\_predicates}(\mathcal{G}, g)$
    \If {$|s|$=$|g|$}
        \State $p_{\text{best}} = p$
        \State \textbf{break}
    \EndIf

    \If {$|visited|=N$}
        \State \textbf{break}
    \EndIf
    
    \If {$|s|> n_{\text{best}}$}
        \State $p_{\text{best}} = p$
        \State $n_{\text{best}}=|s|$
    \EndIf
    
    \State $\mathcal{A}_{\text{precond}} = \text{[]}$
    \For{$a \in \mathcal{A}$}
        \If {$a$ contains any of the target objects}
            \If {$P(\mathcal{G}, a)$}
                \State $\mathcal{A}_{\text{precond}}\text{.append(}a\text{)}$
            \EndIf
        \EndIf
    \EndFor
    \State $\mathcal{A}_{\text{LLM}} = \text{LLM\_Filter(}\mathcal{G}, \mathcal{A}_{\text{precond}}, g\text{)}$
    \For{$a \in \mathcal{A}_{\text{LLM}}$}
        \State $\mathcal{G}^{\text{new}} = E(\mathcal{G}, a)$
        \State $p^{\text{new}} = p + \text{[}a\text{]}$
        \State $s^{\text{new}} = \text{get\_satisfied\_predicates(}\mathcal{G}^{\text{new}}, g\text{)}$
        \State $f^{\text{new}} = |p^{\text{new}}| + (|g|-|s^{\text{new}}|) * k$
        \State $\mathcal{F}.\text{push((}\mathcal{G}^{\text{new}}, p^{\text{new}}, f^{\text{new}}\text{))}$
    \EndFor
    
\EndWhile
\State \Return $p_{\text{best}}$
\end{algorithmic}
\end{algorithm*}

\section{Details of The Finetuned Agent}
For TESS-Qwen, we use a training set of 3k task instances for finetuning the translation, exploration, and filtering modules. For each module, the training sample is structurally similar to the query format used in the prompts described in Section~\ref{sec:supp-ZS}, though the input here does not include in-context learning examples or detailed problem descriptions. Below, we describe the procedures for constructing the training data.

For the \textbf{translation} problem, each training sample corresponds to a single task instance. The input is the task instruction, while the groundtruth output is the list of goal conditions expressed in the form of grounded predicates.

For the \textbf{exploration} problem, we employ an expert exploration policy augmented with random perturbations to collect training samples on the training tasks. Specifically, at each time step, for a target object that has not yet been observed, the policy selects either an optimal exploration direction (\textit{i.e.}, the room or container where the target object is located, according to the full SG of the environment) with a probability of 50\%, or a random direction with the remaining 50\%, and continues this process until all target objects have been observed. At each time step, we record three elements: the aggregated partial SG in the agent's memory, the target object to be explored, and the optimal exploration direction. The first two serve as model input during training, while the third is used as supervision. This approach yields approximately 12k training samples.

For finetuning LLM-based \textbf{filtering}, we employ a similar random-expert policy as the exploration stage for data collection. Given a memory SG (where all the target objects are observed) as the initial state, the agent identifies a set of ``good'' actions at each timestep. It selects a ``good'' action from this set with a probability of 80\%; or selects an action outside this set with the remaining 20\% probability. For each step, we record the current SG and the complete set of candidate actions (filtered by target objects and preconditions) as input, and use the ``good'' action set as supervision. This process yields approximately 100k training samples. The method for determining the ``good'' actions follows a similar procedure to the expert trajectory construction described in Section~\ref{sec:supp-dataGen}: we examine all currently unsatisfied goal conditions, use a heuristic strategy to generate a subplan for each of them, and collect the first action from each subplan.

For each of the three training processes, we finetune a Qwen3‑8B base model using LoRA~\cite{lora} (rank=8, $\alpha$=256). Optimization is performed with AdamW, and a learning rate of 5e-5 with cosine decay. The batch size is 32 (one sample per GPU on 4 GPUs with 8‑step gradient accumulation), the gradient is clipped to a norm of 1.0, and the input sequences are truncated to 2,048 tokens. We train each model for 3 epochs on 4 NVIDIA RTX 4090 GPUs. The three finetuning stages require approximately 0.3h, 1.2h, and 10h, respectively.

\section{Details of the Compared Baselines}
Here we describe the implementation details of the reproduced baselines in the main experiment (Table~\ref{tab:main} in the main text).
\paragraph{CoT+ZSP} ZSP~\cite{LMZSP} is originally a zero-shot method for ungrounded task planning that ignores the environment information. We augment its prompt with a textual description of the initial SG to enable the model to generate plans grounded in specific environments. 
The prompt also includes the definitions and formats of 13 atomic actions, along with several environment constraints (\textit{e.g.}, the agent can hold no more than two objects simultaneously). To facilitate in-context learning, we manually craft five examples with the task instruction, the environment information and the complete plan. Each example also includes an explicit reasoning process to encourage the model’s chain-of-thought (CoT)~\cite{CoT} behaviors. Note that this baseline requires access to the full SG as the environment context. With the partial observation, the model will lack sufficient information to infer the locations of the unobserved target objects, making it difficult to generate a correct plan.

In the original ZSP framework, a post-processing step is applied for each generated action to map it to a valid action using semantic similarity. We remove this post-processing and instead let the model generate the full plan directly. This modification is motivated by two considerations: first, when the environment contains hundreds of objects, the space of valid actions becomes extremely large; second, we observe that when demonstrations and action formats are provided in the prompt, the model generally produces syntactically well-formed actions. 

\paragraph{ReAct} \cite{ReAct} is a prompting technique designed for zero-shot multi-step decision problems. It guides the LLM to generate interleaved sequences of thoughts and actions through in-context learning, thereby enhancing its reasoning capabilities. We implement a ReAct-style policy on VH. At each time step, the model takes as input the task instruction, action history, and the agent's current memory SG (\textit{i.e.}, the aggregation of historical partial observations). It generates an analysis about its goal and situation, followed by the action for the next step. The prompt includes the same meta-information as used in ZSP, along with 5 examples demonstrating the generation of thought and action. The action space is augmented with an additional ``stop'' action, and the episode will also be forced to stop if the number of time steps exceeds 200.

\paragraph{Reflexion}\cite{Reflexion} is a replanning technique that helps the LLM reflect on previous experiences and make better decisions. We add a reflection module to the ReAct agent to implement a Reflexion-style baseline on VH. Specifically, after the agent executes each action, we retrieve feedback from the environment. If the action fails, we provide the LLM with the action history and the failure reason (\textit{e.g.}, ``agent is not close to \verb+<radio> (219)+ when executing \verb+[SWITCHON] <radio> (219)+''), prompting it to reflect on the failed plan and generate advices for the next trial. We then reset the environment to its initial state and incorporate the failed plan along with the corresponding revision suggestions into the prompt, initiating a new ReAct rollout. This reflection process is performed for up to 5 rounds. Note that the access to environment feedback and the ability to reset the environment are assumed not available in TESS and other baselines.

\paragraph{ProgPrompt}~\cite{PROGPROMPT} is a code-based planning method. It originally takes the task name and a list of available objects as input, and generates a program which can be parsed into a sequence of actions. The program includes several ``assert-else'' statements, which invoke the LLM to analyze the current environment state, assess whether the assertion is satisfied, and execute the else branch if the assertion does not hold. However, we observe that such queries are unnecessary in our environment, as the assertion can be directly verified by checking the SG. Therefore, we remove these additional LLM calls and replace the ``assert-else'' statements with more straightforward ``if-else''.
In addition, we introduce ``for'' loops to enhance the expressiveness of the programs generated by ProgPrompt. These loops allow the agent to sequentially check a list of objects and terminate once a specified condition is met (by ``break''). We manually write 5 example programs containing ``for'' and ``if-else'' structures for in-context learning. 
Importantly, these programs are written in standard Python syntax, so the LLM's outputs can be directly saved as Python files. This allows us to call the generated functions for plan execution without requiring an additional step to parse the program text into a sequence of actions.

\paragraph{FLTRNN}~\cite{FLTRNN} is also a code-based planning method. It first leverages an LLM to decompose the task into a series of subtasks. Then for each subtask, it queries the LLM to generate a plan in the form of pseudocode. The prompt includes the environment information (full SG), some rules of the simulator (\textit{e.g.}, the agent can hold no more than two objects simultaneously), the global task instruction and the local subtask, and some demonstrations. We craft 5 demos for the task decomposition stage, and 5 demos for the subtask planning stage. We execute the plan for each subtask once it is generated, and provide the updated environment SG in the prompt for the next subtask.

\paragraph{TreePlanner}~\cite{TreePlanner} is a replanning strategy for zero-shot task planning. It first generates $N$ plans according to the instruction and a list of available objects in the scene. These plans are merged into a tree structure, and a grounded decision approach is proposed to traverse this tree for a successful plan, under the guidance of an LLM. We note that the purpose of performing grounded decision over the plan tree is to reduce the number of interactions with the environment. However, this process is relatively complex and incurs additional LLM queries. On the other hand, the set of $N$ sampled plans has already determined the upper bound of TreePlanner's performance. Therefore, we adopt a simplified TP-upper-bound baseline, in which the $N$ plans are executed sequentially, and the best-performing one is recorded for evaluation. Each plan is generated in a manner similar to ZSP+CoT, except that the full SG in the prompt is replaced with an object list. $N$ is set to 50 in our experiments.

\paragraph{LID}~\cite{LMIDM} is a finetuning-based action policy. It takes the instruction, the environment observation, and the action history as input, and outputs an atomic action or ``stop''. We maintain a memory SG for the LID agent using a method similar to the exploration stage of TESS, and this aggregated memory is verbalized and used as the observation. The 3k training tasks with expert trajectories are used as training data, and each step in the trajectory corresponds to one training sample. We finetune a Qwen3-8B for this baseline.

\paragraph{EquilibriumPlanner}~\cite{ESM} is a finetuning-based method for plan generation. It proposes a supervised training pipeline that equips the agent with self-refining capabilities through equilibrium sequence modeling. Specifically, it trains the model to refine an equilibrium plan to an expert plan. During inference, the model can perform multiple rounds of plan refinement by fixed-point iteration. EquilibriumPlanner allows the incorporation of environment feedbacks during the refinement process, in order to facilitate more effective plan improvements. However, we do not implement this in our reproduction, as acquiring feedbacks during training would require loading the full SG for all tasks and constructing corresponding simulators. Given the large number of training tasks and the complexity of the environments, this would incur substantial memory overhead and increase the time for data preparation. Therefore, our reproduced model only takes as input the task instruction, the environment information (a list of available objects in the scene), and the equilibrium plan. We use LLaMA3-8B as the base model and follow the finetuning procedure in the original paper.

From the baseline results presented in Table~\ref{tab:main} of the main text, we have the following findings:
(1) Providing the agent with the full SG as input leads to a clear performance advantage over using a partial SG or an object list, as it eliminates the need for exploration in unknown regions;
(2) Incorporating reflection and refinement mechanisms enhances the performance of LLMs (Reflexion, EquilibriumPlanner), particularly on unseen splits;
(3) Program-based plan generation (FLTRNN vs. CoT+ZSP) demonstrates certain advantages on unseen splits;
(4) Allowing the agent to attempt multiple plans (TP vs. ProgPrompt) significantly boosts overall performance;
(5) Fine-tuning a smaller open-source model on a limited amount of data can outperform a larger proprietary model given the same environmental input (LID vs. ReAct, EquilibriumPlanner vs. ProgPrompt), although generalization to unseen splits remains a challenging issue.

In Table~\ref{tab:AW} of the main text, the baselines on ALFWorld are implemented following their open source codes. To ensure a fair comparison, we made minimal modifications to guarantee that (1) all methods use the GPT-4.1-mini API, and (2) the maximum number of action steps is set to 100. 
It is worth noting that some baselines require assumptions that are not adopted by TESS. For example, Reflexion~\cite{Reflexion} allows multiple trials within the environment; Expel~\cite{Expel} introduces an additional experience gathering phase; RAP~\cite{RAP_AW} performs retrieval-augmented inference based on prior task plans.

\section{More Experimental Results}
\begin{figure}[]
\begin{minipage}{0.48\linewidth}
        \centering
		\includegraphics[width=0.98\columnwidth]{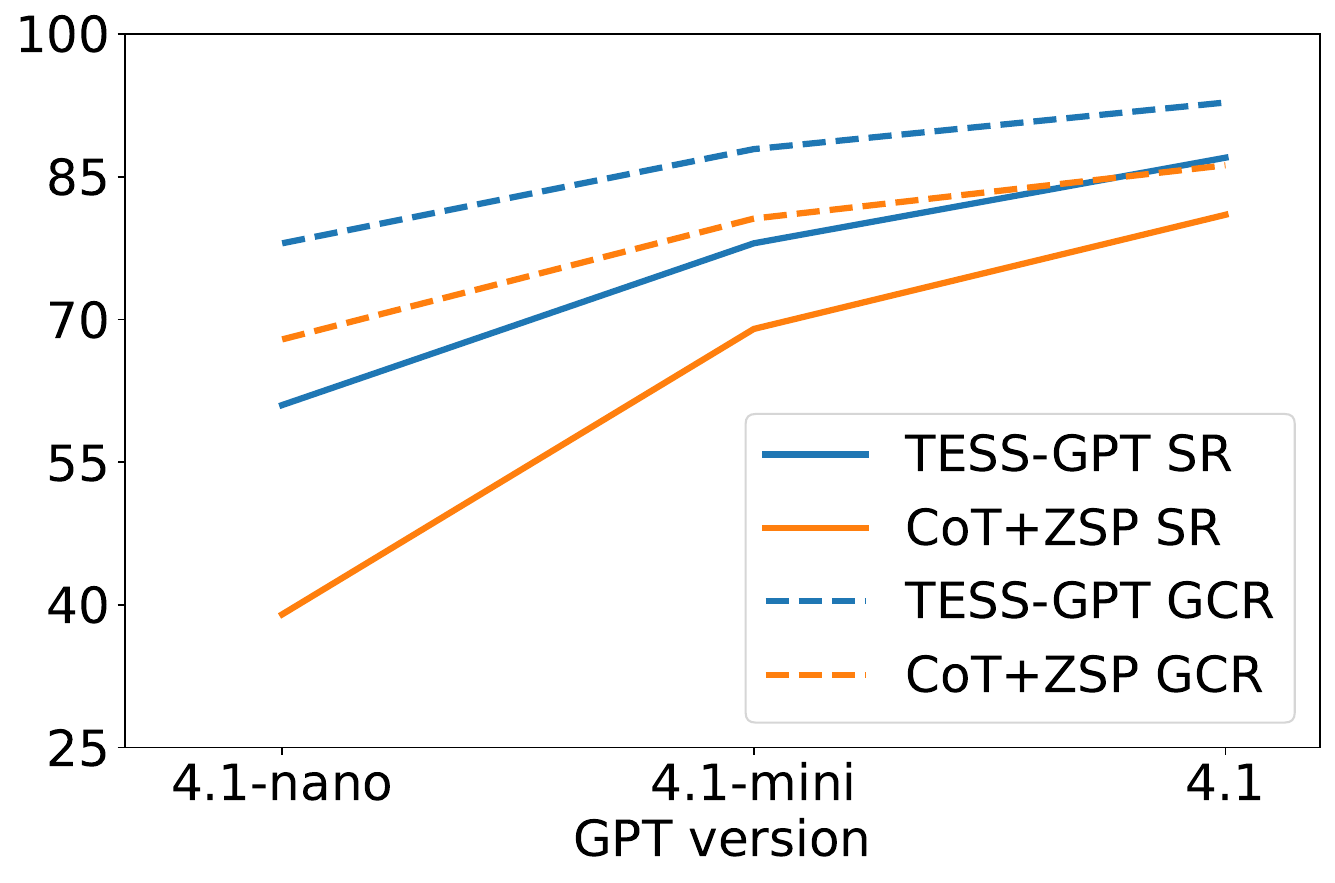}
		\caption{Performance with different versions of GPT.}\label{fig:supp-scale_zs}
\end{minipage}
\begin{minipage}{0.48\linewidth}
        \centering
		\includegraphics[width=0.98\columnwidth]{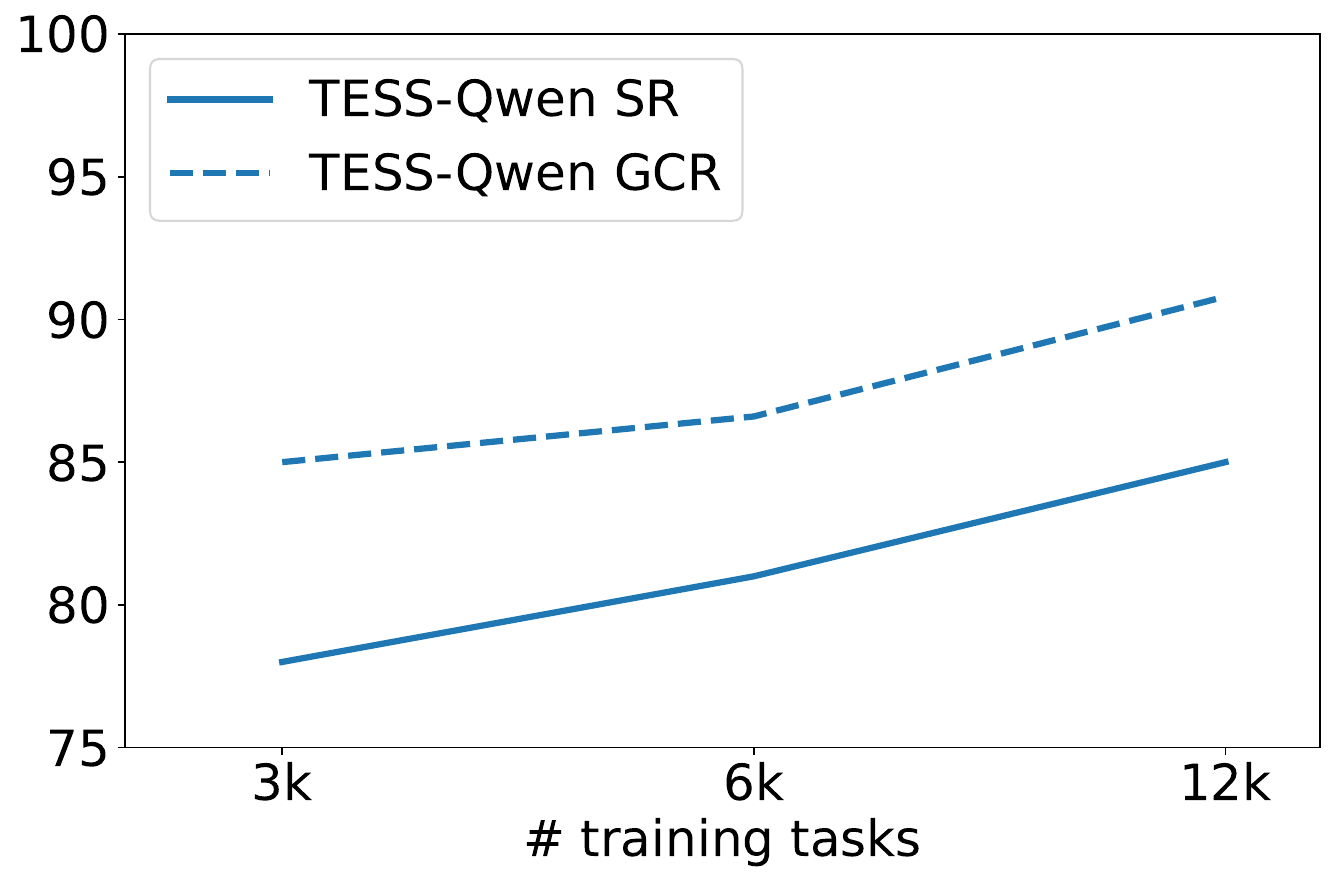}
		\caption{Performance with different size of training set.}\label{fig:supp-scale_ft}
\end{minipage}

\end{figure}

\subsection{Additional Ablation Studies}
In this section we present some fine-grained ablation experiments on the design of the proposed pipeline. Unless stated otherwise, all the results reported in this section are obtained on a tiny test split of 100 tasks, due to the constraints in time and budget.

First, for the translation module, we compare our predicate-based goal state checking with querying LLMs to determine the termination. As shown in table~\ref{tab:supp-trans}, we show the accuracy for LLM-translation, the success rate for LLM-goal-checking, and the average number of tokens needed for each task in both methods. The success rate is computed as follows: for each task, we execute the expert plan step by step. After each action execution, the LLM is prompted to assess whether the current state satisfies the task instruction. A task is considered successful only if the LLM returns ``No'' at every intermediate step, and outputs ``Yes'' after the final action. In summary, the translation-based goal state checking is significantly more reliable and more efficient than directly querying the LLM to assess the state.
\begin{table}
  
  \centering
  \small
  \scalebox{0.9}{
  \begin{tabular}{lcccc}
    \toprule
  & ACC/SR & \#call & \# token in & \# token out \\
  \midrule
  Translation & 0.94 & 1 & 639 & 24 \\
  Goal Checking & 0.67 & 10 & 124,838 & 36 \\
    \bottomrule
  \end{tabular} 
}
  \caption{Comparison between LLM-based translation and LLM-based goal state checking.} 
  \label{tab:supp-trans}
  \vspace{-3mm}
\end{table}

Second, for the simulation module, we also intend to conduct an experiment similar to Table~\ref{tab:supp-trans} to compare the performance of our rule-based world model with directly querying an LLM to simulate state transitions. However, we found that the latter approach incurs unacceptable computational overhead, as both the input and output involve a textualized SG that contains a large number of objects. Moreover, parsing the generated textual descriptions into a valid SG requires additional engineering effort. Therefore, we opt to omit this comparison, as the advantages of rule-based world model—in terms of both efficiency and interpretability—are evident enough.

Next, for the search module, we have already shown the advantage of replacing BFS with A* search in Table~\ref{tab:search} of the main text.
Here we further present a quantitative evaluation of the effects of each filtering operation in Table~\ref{tab:supp-filter}. We consider the first search step in each episode, and record the size of the full action space $\mathcal{A}$, the action space $\mathcal{A}_{\text{obj}}$ after filtering by target objects, the action space $\mathcal{A}_{\text{precond}}$ after filtering by preconditions, and the action space $\mathcal{A}_{\text{LLM}}$ after filtering by the LLM (GPT-4.1-mini). It can be seen that the number of possible actions have been gradually, significantly reduced by these filters, enabling an efficient search process that incorporates commonsense knowledge.

\begin{table}
  
  \centering
  \scalebox{0.9}{
  \begin{tabular}{cccc}
    \toprule
  $\left|\mathcal{A}\right|$ & $\left|\mathcal{A}_{\text{obj}}\right|$ & $\left|\mathcal{A}_{\text{precond}}\right|$ & $\left|\mathcal{A}_{\text{LLM}}\right|$ \\
  \midrule
  175,800 & 83 & 4 & 2 \\
  \bottomrule
  \end{tabular}
  }
  \caption{The average size of the filtered action spaces.} 
  \label{tab:supp-filter}
  \vspace{-3mm}
\end{table}

Finally, we investigate how performance varies with respect to model size and the amount of training data. For TESS-GPT, we consider the use of different backbones for the translation and filtering modules.\footnote{We keep using the lightweight GPT-4o-mini model for the exploration module, as exploration is a relatively straightforward task and it only affects the length of the generated plan rather than the correctness. The simulation module is fixed to GPT-4.1, since it involves only a minimal number of input and output tokens throughout the experiment.} As in Figure~\ref{fig:supp-scale_zs}, TESS consistently outperforms the CoT+ZSP baseline under all of the three settings, demonstrating the generality of the proposed method across different computational budgets.
For TESS-Qwen, we try to enlarge the scale of training data for the three finetuning processes. In Figure~\ref{fig:supp-scale_ft}, the agent consistently benefits from the increased amount of training data, indicating both the scalability of our agent and the effectiveness of the data collection procedure.

\subsection{Qualitative Results}

\begin{figure*}[h!]
\centering
\includegraphics[width=1.8\columnwidth]{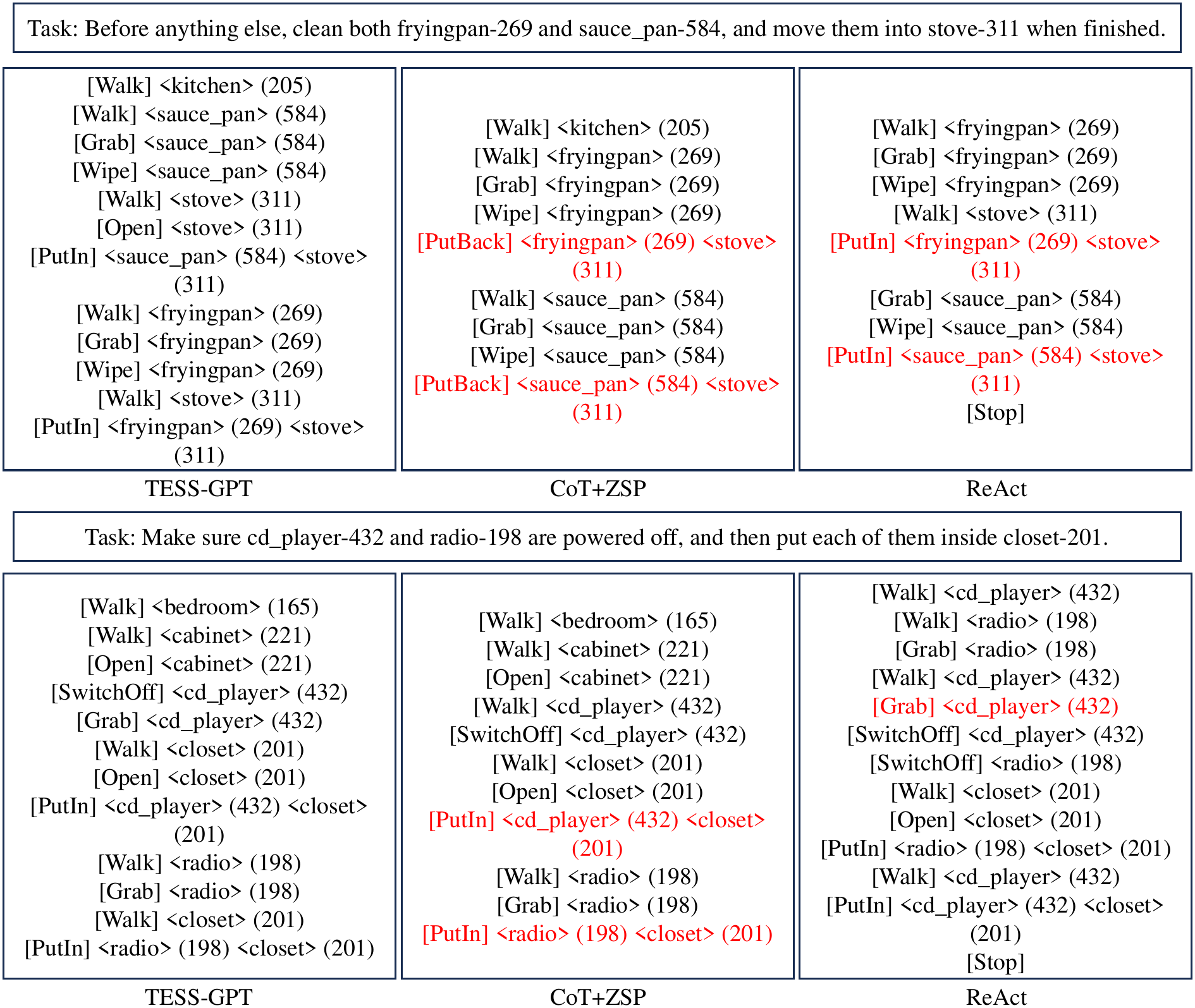}
\caption{Two qualitative examples of the generated plans. Our TESS-GPT agent generates successful plans for both tasks while the CoT+ZSP and ReAct baselines fail. The steps that lead to failures are marked in red.}\label{fig:supp-vis}
\end{figure*}

\begin{figure*}[h!]
\centering
\includegraphics[width=1.8\columnwidth]{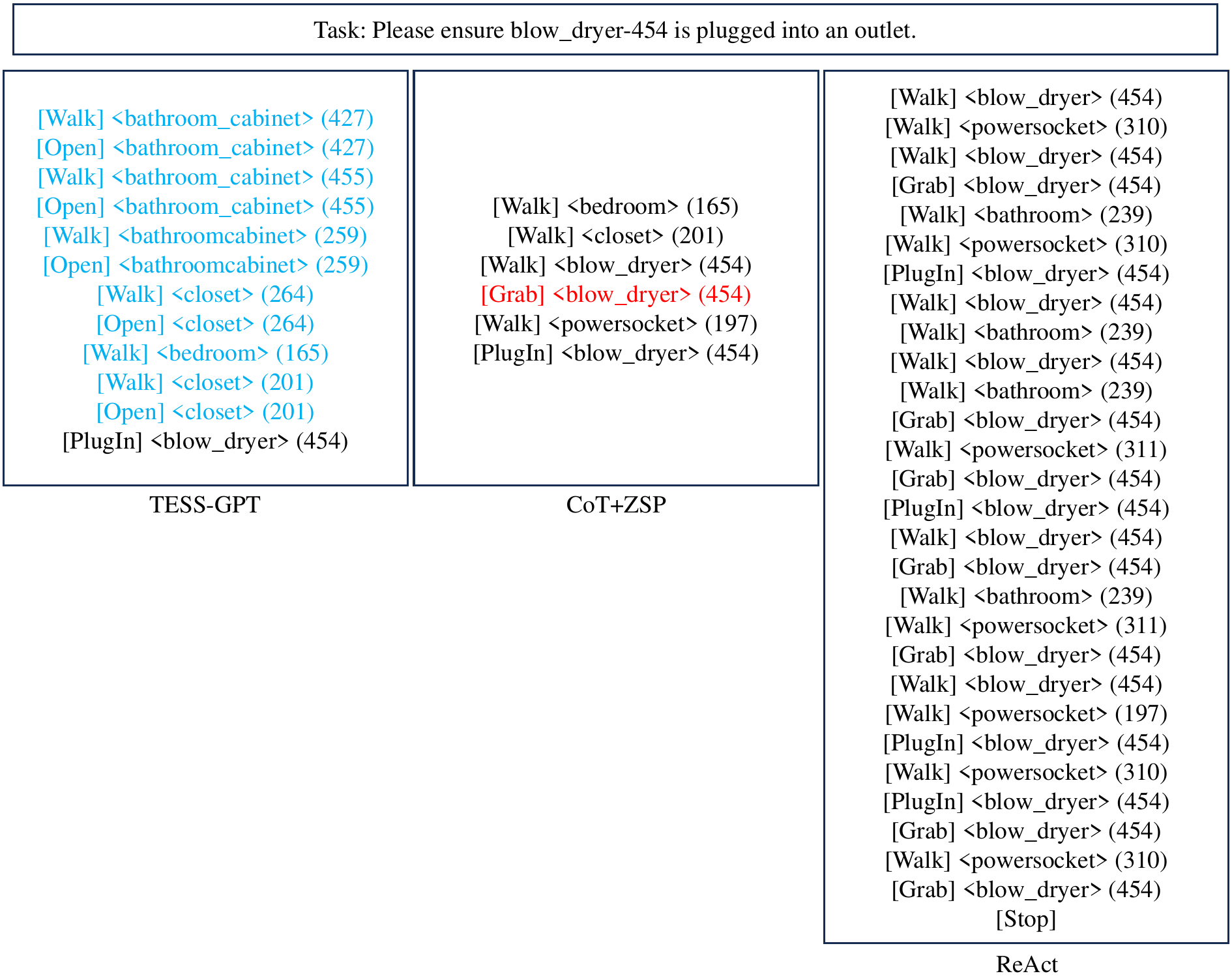}
\caption{The example of an exploration-intensive task, where the agent has to find the target blow\_dryer before manipulating it. For TESS-GPT, the actions for exploration are marked in blue. For CoT+ZSP, the full SG is provided so the agent can go directly to the closet where the blow\_dryer is located. However, it forgets to open the closet before grabbing the blow\_dryer. The ReAct agent keeps trying repeated actions without making meaningful progress and never manages to locate the target object throughout the episode. The results suggest that incorporating an explicit exploration process can be beneficial for task planning.}\label{fig:supp-vis2}
\end{figure*}

Some visualizations of the generated plans are shown in Figure~\ref{fig:supp-vis} and \ref{fig:supp-vis2}. The baseline methods frequently encounter several types of failures, including the omission of critical steps (\textit{e.g.}, placing an object into the stove without opening the stove, grabbing an object in the cabinet without opening the cabinet), misinterpretation of instructions (\textit{e.g.}, placing an object on the stove rather than inside it), and insufficient exploration (\textit{e.g.}, failing to find the target object). Thanks to the explicit translation, exploration, and world modeling, our TESS agent can effectively handle these problems.

\section{Error Analysis}

\subsection{Overview}

We begin the error analysis by attributing the failure cases to different modules. As in Table~\ref{tab:supp-gt}, replacing the translation module and the simulation module with their groundtruth counterparts leads to success rate improvements of 2.0\% and 10.5\%, respectively. This indicates that the simulation module has greater space for improvement. All remaining failures are caused by the search stage.
If we further relax the limit on the number of search steps, the success rate eventually reaches 100\%, validating the theoretical performance guarantee of classical search-based methods.
In the following, we provide a detailed analysis of the sources of errors associated with each module. It is worth noting that the ability to conduct fine-grained analysis on an LLM's performance across different aspects of the task planning problem is, in itself, one of the advantages of the modular TESS framework.
\begin{table}
  \centering
  \small
  \scalebox{0.9}{
  \begin{tabular}{llccc}
  \toprule
  Translation & Simulation & \# max search & SR & GCR \\
  \midrule
  GPT-4.1-mini & GPT-4.1 & 500 & 80.0 & 88.4 \\
  GT & GPT-4.1 & 500 &  82.0 & 89.7   \\
  GT & GT & 500 & 92.5 & 97.0 \\
  GT & GT & $\infty$ & 100.0 & 100.0  \\
  \bottomrule
  \end{tabular}
  }
  \caption{Error analysis by replacing the translation/simulation module with the corresponding groundtruth, and enlarging the budget for search steps. 
  }  \label{tab:supp-gt}
\end{table}

\subsection{Translation Error}

In Table~\ref{tab:trans} of the main text, we report that the translation module of TESS-GPT achieves an accuracy of 96.0\%. To further understand the remaining 4\% of errors (a total of 40 tasks), we conduct a manual inspection and categorize the errors into the following types.
(1) Non-critical redundant predicates. For example, given the instruction “Please go to dishwasher-103 and close it so nothing is left open”, the translation output is \verb+(close_to, agent-1, dishwasher-103)+ and \verb+(CLOSED, disherwasher-103)+. Here the predicate of ``close\_to'' is an unnecessary requirement. However, even with this redundant predicate, the agent is still able to generate a valid plan to complete the original task.
(2) Critical redundant predicates. Consider an instruction like ``Grab A1 as well as A2 and make sure they are both inside B.'' The LLM fails to distinguish between the intermediate state and the final state, and introduces redundant predicates such as \verb+(holding, agent-1, A1)+ and \verb+(holding, agent-1, A2)+, which are incompatible with the correct final state where A1 and A2 are inside B. This misinterpretation leads to a conflict that prevents the search algorithm from reaching a valid terminal state, thus making the planning process failed.
(3) Incorrect predicate types. 
Consider the instruction ``Move bag-514 so it sits on shoe\_rack-512.'' The LLM's output is \verb+(sitting_on, bag-514, shoe_rack-512)+, rather than the correct answer \verb+(on, bag-514, shoe_rack-512)+. It fails to recognize that the phrase ``sits on'' in the instruction is a figurative way of expressing an \verb+on+ relation between objects, whereas the \verb+sitting_on+ relation can only occur between an agent and an object. Errors of this kind may lead to failures in finding a plan that satisfies the translated predicates, or result in a plan that does not fulfill the groundtruth goal conditions.

In future work, we plan to address issue (2) by analyzing the compatibility among the translated predicates, and address issue (3) by introducing a reflection and goal refinement mechanism after the generation and execution of the plan.

\subsection{Exploration Effectiveness}

The exploration stage has no influence on the task success rate, while it only affects the efficiency of task completion. In Table~\ref{tab:exp} of the main text, we present the number of environment interactions required by TESS during the exploration stage. Additionally, the example shown in Figure~\ref{fig:supp-vis2} qualitatively illustrates the rationality of the exploration steps.
We also note that the cost of exploration can be further reduced if the agent is performing a sequence of tasks in the same environment, where its memory SG can be inherited across episodes. This is a meaningful setting for robots deployed in a household.

\subsection{Simulation Error}

We conduct a manual inspection of the precondition and effect functions generated by the LLM and find that they accurately simulate changes in the environment for most cases. However, they exhibit oversight in handling certain edge cases. Typical overlooked constraints include:
(1) When the agent is holding objects in both hands, it should be unable to operate other objects. 
(2) Objects such as the dishwasher cannot be opened while they are running. 
(3) For objects that have a plug, they cannot be switched on unless they are already plugged in. 

In future work, we plan to develop an interactive simulation module. After generating the precondition and effect functions, the agent will execute either expert plans or random actions in the environment, and the resulting state transitions will be compared against those predicted by the simulator. When discrepancies arise, we will manually revise the specific functions responsible for the errors, or feed the relevant information back into the LLM to enable self-reflection and correction. This iterative process allows the learned world model to be progressively improved with accumulated experiences.

\subsection{Search Failure}

Given accurate grounded predicates and action transitions, BFS is guaranteed to find a correct and optimal (in terms of the number of action steps) plan. Replacing BFS with A* sacrifices a certain degree of optimality in exchange for improved efficiency, while still ensuring the correctness of the resulting plan. In addition, applying filters based on the target objects and action preconditions does not affect the success rate.
Therefore, failures in the search stage can be attributed to two sources:
(1) The LLM-based filter mistakenly removes some correct and necessary actions, making it impossible for the algorithm to find a valid path to the goal state;
(2) The search proceeds normally but is forced to terminate upon reaching the search step limit.

\begin{table*}[t]
\centering
\caption{Results on ALFWorld (visual observation).}
\label{tab:ex_AW}
\scalebox{0.8}{
\begin{tabular}{lccccccc}
\toprule
Method & Avg & Pick & Clean & Heat & Cool & Look & Pick2 \\
\midrule
GPT-4o \cite{SEEA-R1} 
& 24 & 44 & 22 & 29 & 27 & 7 & \textbf{23} \\
TESS 
& \textbf{43} & \textbf{71} & \textbf{42} & \textbf{39} 
& \textbf{62} & \textbf{22} & 12 \\
\bottomrule
\end{tabular}
}
\end{table*}

\begin{table*}[t]
\centering
\caption{Results on EmbodiedBench-ALFRED.}
\label{tab:ex_EB}
\scalebox{0.8}{
\begin{tabular}{lccccccc}
\toprule
Method & Avg & Base & Common & Complex & Visual & Spatial & Long \\
\midrule
GPT-4o \cite{EmbodiedBench}
& 56.3 & 64 & 54 & \textbf{68} & \textbf{46} & \textbf{52} & 54 \\
TESS 
& \textbf{59.3} & \textbf{74} & \textbf{66} & \textbf{68} 
& 34 & 42 & \textbf{72} \\
\bottomrule
\end{tabular}
}
\end{table*}

We hypothesize that failures due to (1) are relatively rare, as there are typically multiple valid paths between two environment states. Removing a few correct actions is unlikely to make the search space disconnected. The influence of (2) can be observed in Table~\ref{tab:supp-gt}, which shows that a search budget of 500 steps is sufficient for solving the majority (92.5\%) of tasks. For the most challenging cases (those with expert trajectories longer than 30 steps), increasing the search limit enables the planner to successfully complete all tasks, which can be seemed as a test-time scaling technique. Another promising research direction is to first decompose the task into a sequence of subtasks that can be executed in order, and then perform a significantly shallower search for each individual subtask.

\section{Extensions}
Our main experiments are conducted in a text environment. However, TESS does not necessarily require a pre-parsed structured scene graph as input, and it is able to take unstructured RGB images as observations. For this setting, we slightly extend the exploration module as follows: after each exploration step, a VLM examines the resulting image to determine whether the target object is visible and updates the agent’s memory scene graph accordingly. 
We implement this extended version of TESS on ALFWorld (visual observations)~\cite{AW} and EmbodiedBench-ALFRED~\cite{EmbodiedBench}. We use the GPT-4o results reported in previous works~\cite{SEEA-R1,EmbodiedBench} as baselines and adopt the same VLM backbone and evaluation protocol to ensure comparability. The results are shown in Tables~\ref{tab:ex_AW} and\ref{tab:ex_EB}.

On ALFWorld, TESS substantially outperforms the baseline, indicating that its benefits are retained when structured symbolic observations are unavailable. On EmbodiedBench, TESS also achieves superior average performance, and its advantage is particularly pronounced on the ``long horizon'' split, suggesting that LLM-guided search is beneficial for tasks requiring longer and more complex action sequences. We also observe that TESS performs less favorably on the ``visual appearance'' and ``spatial relationship'' splits. Many tasks in these splits contain referring expressions defined by visual attributes or spatial positions, and resolving the intended target therefore requires jointly reasoning over the instruction and the visual observation. Currently, our translation module parses the task goal primarily from the textual instruction and may therefore fail to identify the target object accurately. We view the current experiment as an initial demonstration for TESS's multi-modal extension. A promising future direction is to integrate the translation module with the exploration process, allowing instruction interpretation and visual grounding to be updated jointly as new observations are acquired.